%% file: ecai.tex
\documentclass[camera-ready]{ecai}
\usepackage{times}
\usepackage{soul}
\usepackage{url}
\usepackage[hidelinks]{hyperref}
\usepackage[utf8]{inputenc}
\usepackage[small]{caption}
\usepackage{graphicx}
\usepackage{amsmath}
\usepackage{amsthm}
\usepackage{booktabs}
\usepackage{adjustbox} 

\usepackage[switch]{lineno}
\paperid{1646}

\newcommand{\shrink}[1]{}

\usepackage{tikz} 
\usepackage{graphicx} 
\usepackage[font=footnotesize,labelfont=bf]{subcaption}
\usepackage[font=footnotesize,labelfont=bf]{caption}
\usepackage[ruled,vlined]{algorithm2e}

\usetikzlibrary{calc}
\usetikzlibrary{arrows, automata}
\usetikzlibrary{shapes}
\usetikzlibrary{positioning, quotes}

\usepackage[utf8]{inputenc} 
\usepackage[T1]{fontenc}   
\usepackage{url}            
\usepackage{booktabs}       
\usepackage{amsmath,amssymb, amsfonts}       
\usepackage[ruled,vlined]{algorithm2e}
\SetCommentSty{commentstyle}
\SetKwInOut{Input}{input}
\SetKwInOut{Output}{output}
\newcommand{\hrulealg}[0]{\vspace{1mm} \hrule}
\usepackage{calc}
\usepackage{nicefrac}       
\usepackage{microtype}      
\usepackage{xcolor}         

\newcommand{\G}{\mathcal{G}}
\newcommand{\M}{\mathcal{M}}
\newcommand{\B}{\mathcal{B}}
\newcommand{\F}{\boldsymbol{F}}
\newcommand{\D}{\mathcal{D}}

\newcommand{\U}{\boldsymbol{U}} 
\newcommand{\V}{\boldsymbol{V}} 
\newcommand{\YY}{\boldsymbol{Y}} 
\newcommand{\XX}{\boldsymbol{X}} 
\newcommand{\xx}{\boldsymbol{x}} 
\newcommand{\btheta}{\boldsymbol{\theta}}
\newcommand{\kk}{\boldsymbol{k}}
\newcommand{\PA}{P\!A}

\newtheorem{example}{Example}

\newtheorem{theorem}{Theorem}
\newtheorem{definition}{Definition}

\begin{document}


\begin{frontmatter}


\title{Estimating Causal Effects from Learned Causal Networks}

\author[A]{\fnms{Anna}~\snm{Raichev}\thanks{Corresponding Author. Email: araichev@uci.edu.}}
\author[B]{\fnms{Jin}~\snm{Tian}} 
\author[A]{\fnms{Alexander}~\snm{Ihler}}
\author[A]{\fnms{Rina}~\snm{Dechter}} 

\address[A]{University of California, Irvine}
\address[B]{Iowa State University}

\begin{abstract}
The standard approach to answering an identifiable causal-effect query (e.g., $P(Y|do(X)$) given a causal diagram and observational data is to first generate an estimand, or probabilistic expression over the observable variables, which is then evaluated using the observational data. In this paper, we propose an alternative paradigm for answering causal-effect queries over discrete observable variables. We propose to instead learn the causal Bayesian network and its confounding latent variables directly from the observational data. 
Then, efficient probabilistic graphical model (PGM) algorithms 
can be applied to the learned model to answer queries.
Perhaps surprisingly, we show that this \emph{model completion} learning approach can be more effective than estimand approaches, particularly for larger models in which the estimand expressions become computationally difficult.
We illustrate our method's potential using a benchmark collection of Bayesian networks and synthetically generated causal models.
\end{abstract}
\end{frontmatter}

\section{Introduction}

Structural Causal Models (SCMs) are a formal framework for reasoning about causal knowledge in the presence of uncertainty \cite{pearlbook}. When the full SCM is available, it is possible to use standard probabilistic inference \cite{DarwicheBook09,DBLP:series/synthesis/2013Dechter} to
directly answer causal queries that evaluate how forcing some variable's assignment $X=x$ affects another variable $Y$, written as $P(Y|do(X=x))$.  
However, in practice the full causal model is rarely available; instead, only a causal diagram -- a directed graph capturing the causal relationships of the underlying SCM -- is assumed to be known. Causal diagrams typically include both \textit{observable variables}, which can be measured from data, and \textit{latent variables}, which are unobservable and for which data cannot be collected. A linchpin of causal reasoning is that given a causal diagram, many causal queries can be uniquely answered using only the observable variables
\cite{DBLP:conf/aaai/TianP02,shpitser2006identification,shpitser2006identificationB,huang2008completeness}
and consequently estimated using observational data.
\nocite{jinthesis,huang2008completeness}
 
The main approach developed in the past 
two decades for answering causal queries under such assumptions is a two-step process which we call \emph{estimand-based causal inference}. 
The first step is to determine if the causal query is \emph{identifiable} -- i.e., uniquely answerable from the model's observational distribution -- and if so, construct an expression (``estimand'') that captures the answer symbolically using probabilities over only observed variables. 
Then, one can use the observed data to estimate the probabilities involved in the estimand expression.
Over the past few years a variety of estimand-based strategies have been developed using different statistical estimation methods 
\cite{DBLP:conf/aaai/Jung0B21,bhattacharya2020semiparametric}\nocite{DBLP:conf/aaai/Jung0B20,DBLP:conf/nips/Jung0B20,DBLP:conf/icml/Jung0B21}.

However, many of these approaches focus on specific, small, causal models, and do not examine the scalability of their approaches or provide significant empirical comparisons.
In larger models, estimand expressions can become large and unwieldy, making them either computationally difficult, hard to estimate accurately, or both.

In this paper, we propose a straightforward yet under-explored alternative: to learn the causal model, including latent dependencies, directly from the observed data.
Although the domains and distributional forms of these latent variables are unknown, recent work \cite{DBLP:conf/icml/Zhang0B22} has shown that, if the visible variables are discrete, any SCM is equivalent to one with discrete latent variables, and gives an upper bound on their domain sizes.
This allows us to apply well-known techniques for learning latent variable models,
such as the Expectation-Maximization (EM) algorithm, to build a
{\em Causal Bayesian Network (CBN)} over the observed and latent variables.  
We can further apply model selection techniques such as the Bayesian Information Criterion (BIC) \cite{koller} to select appropriate domain sizes.
Then, given our learned model, we can use efficient algorithms for reasoning in probabilistic graphical models (PGMs) to answer one, or even many causal queries
\cite{DarwicheBook09,koller}\nocite{DBLP:series/synthesis/2013Dechter,pearlbook}. 

Perhaps surprisingly, we show that this approach is often significantly
more effective than the estimand-based methodology. 
Furthermore, we provide structural conditions that help decide when each approach is likely to be more effective. 
The computational benefit of our approach is tied to the complexity of the causal graph: if the causal graph has low {\em treewidth}, then exact PGM algorithms (e.g., bucket elimination or join-tree scheme \cite{DBLP:journals/ai/Dechter99,DBLP:journals/ai/KaskDLD05}) are efficiently applicable.
As an added benefit if there are multiple causal queries to perform on the same model, since the model is learned only once, the learning time can be amortized over all such queries. 

Our empirical evaluation incorporates a spectrum of models, including not only the small models common in causal effect literature, but also large models based on benchmark Bayesian networks and scalable classes of synthetic models; to the best of our knowledge this is the first such extensive empirical evaluation of causal effect queries.

\paragraph{Contributions.}
This paper presents a new path for answering causal effects queries by learning a Causal Bayesian Network that is consistent with a causal graph and observational data. 
\begin{itemize}
\item We propose to answer causal queries by directly learning the full causal Bayesian network via EM, followed by query processing using traditional PGM algorithms.
\item We provide a first of its kind, empirical evaluation of algorithms for causal effect queries on varied and large synthetic and real benchmarks.
%
\item We show empirically that our proposed learning approach gives more accurate estimates than the estimand-based alternatives.
\end{itemize}
In settings with high-dimensional estimand expressions but low treewidth causal models, our approach allows more accurate estimates by retaining more information from the causal graph, and allowing variance reduction without impacting computational tractability.


The rest of the paper is organized as follows. Section \ref{background} provides background on related work and information and definitions, Section \ref{method} outlines the main idea of our approach, Section \ref{eval} provides empirical evaluation, and we conclude in Section \ref{conclusion}.

\begin{figure}[t]
     \centering
          \begin{subfigure}[b]{\columnwidth}
           \centering
           \resizebox{.6\linewidth}{!}{\input{tikzfigures/chain_7.tikz}}  
           \caption{Causal diagram of a chain model with 7 observable and 3 latent variables. Dashed bidirected edges represent latent variables.\\}
           \label{fig:chain_scm}
      \end{subfigure}
      \hfill
     \begin{subfigure}[b]{\columnwidth}
           \centering
           \resizebox{.6\linewidth}{!}{\input{tikzfigures/chain_bn.tikz}}  
           \caption{The CBN for the model (a), showing latent variables explicitly.\\}
           \label{fig:before_do}
      \end{subfigure}\\
      \vspace{5mm}
          \begin{subfigure}[b]{\columnwidth}
           \centering
           \resizebox{.6\linewidth}{!}{\input{tikzfigures/chain_trunc.tikz}}  
           \caption{The truncated causal diagram  after intervention $do(V_0)$.\\}
           \label{fig:after_do}
      \end{subfigure}\\
    \hfill 
      \caption{Illustration of a causal diagram, CBN, and truncated CBN for evaluating a causal effect. Blue variables are intervened on and red variables are the outcome variables corresponding to the query $P(Y \mid do (X))$}
     \vspace{2em}
     \label{fig:small_chain}
 \end{figure}
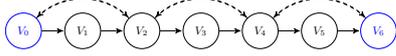
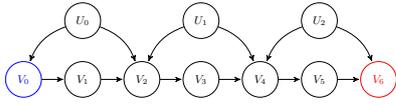
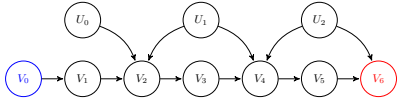

\section{Background}\label{background}
%
We use capital letters $X$ or $V$ to
represent variables, and lower case $x$ or $v$ for their values.
Bold uppercase ${\XX}$ 
denotes a collection of variables, with
$|{\XX}|$ its cardinality and $\D({\XX})$ their joint domain, while $\xx$ indicates an assignment in that joint domain.
We denote by $P({\XX})$ the joint probability 
distribution over ${\XX}$ and $P({\XX=\xx})$ the probability of ${\XX}$ taking configuration $\xx$ under this distribution, which we abbreviate  $P(\xx)$.
Similarly, $P({\YY} | {\XX})$ represents the set of
conditional distributions $P({\YY} \mid {\XX} = \xx)$, $\forall \xx$.
In the definitions ahead we will use $\V$ and $\U$ to stand for arbitrary variables in  models while use $X$ and $Y$ to refer specifically to variables that appear in queries.

\begin{definition}[Structural Causal Model] \label{def:scm}
A structural causal model (SCM) \cite{pearlbook} is a $4$-tuple $\M = \langle \U, \V, \F, P(\U) \rangle$ where: (1) $\U
= \{U_1, U_2, ..., U_k\}$ is a set of exogenous latent variables whose values are affected by factors outside the model;
   (2) $\V =
\{V_1, V_2, ..., V_n\}$ is a set of endogenous, observable
   variables of interest whose values are determined by other variables in the model;
(3) $\F = \{f_{i}: V_{i} \in \V\}$ is a set of functions $f_{i}$ such that each $f_i$ determines the value $v_{i}$ of $V_{i}$ as a function of $V_{i}$'s causal parents $\PA_{i} \subseteq \U\cup(\V\setminus V_{i})$ so that $f_{i} : \D(\PA_{i}) \rightarrow \D(V_{i})$ and $v_{i} \leftarrow f_{i}(pa_{i})$; 
and (4) $P(\U)$ is a probability distribution over the
latent variables. The latent variables  are assumed to be mutually independent, i.e., $P(\U) = \prod_{U \in \U} P(U)$. 
\end{definition}

\paragraph{Causal diagrams.} An SCM $\M = \langle \U, \V, \F, P(\U) \rangle$  induces a {\em causal diagram} $\G = \langle \V \! \cup \U, E \rangle$, where each node in the graph  maps to a variable in the SCM, and there is an arc from every observed or latent node $X$ in the graph to $V_i$ iff  $X \in \PA_{i}$. Thus $\PA_{i}$ are the parents nodes of node $X$ in $\G$. An SCM whose latent variables connect to at most a single observable variable is referred to as {\em Markovian}, and one whose latent variables connect to at most two observable variables is called {\em Semi-Markovian}. It is known that any SCM can be transformed into an equivalent Semi-Markovian one such that answers to causal queries are preserved \cite{jinthesis}.
In the semi-Markovian case it is common to use a simplified causal diagram called an \emph{Acyclic Directed Mixed Graph} (ADMG), 
which omits any latent variables with a single child, and 
replaces any latent variable with two children with a bidirectional dashed arc between the children, so that latent variables are no longer explicitly shown (see, e.g., Figure
\ref{fig:small_chain}). 

The SCM $\M$ also induces a {\bf Causal Bayesian Network (CBN)} $\B = \langle \G, \mathcal{P} \rangle$ specified by $\M$'s causal diagram $\G = \langle \V \! \cup \U, E \rangle$ along with its associated conditional probability distributions $\mathcal{P} = \{P(V_{i}|\PA_{i}),  P(U_j)\}$. 
The distribution $P(\V, \U)$ factors according to the causal diagram:
\begin{align}\label{eq-cbn}
 P(\V, \U) =  \prod_{V_i \in \V} P(V_{i}|\PA_{i}) \cdot \prod_{U_j \in \U} P(U_j).
\end{align}
The \emph{observational distribution},  $P(\V)$, is given by 
\begin{equation}\label{eq-obs}
P(\V) = \sum_{\U}  P(\V, \U).
\end{equation}

\paragraph{Causal effect and the truncation formula.} 
An external intervention, forcing variables $\XX \subseteq  \V$ to take on value $\xx$, is modeled by replacing the mechanism for each $X \in \XX$  with the function $X=x$. This is formally represented by the do-operator $do(\XX=\xx)$. 
Thus the interventional distribution of an SCM $\M$ when applying $do(\XX)$ is defined by,
\begin{equation} \label{eq-intervene}
P(\V\setminus\XX,\U \mid do(\XX=\xx) ) = \prod_{V_j \notin {\XX}} P(V_{j}|\PA_{j}) \cdot P(\U) \bigg|_{\XX=\xx} 
\end{equation}
i.e., it is obtained from Eq.~\eqref{eq-cbn} by truncating from $\M$ the factors corresponding to the variables in $\XX$ and setting $\XX=\xx$. 
The effect of $do(\XX)$ on a variable $Y$, denoted $P(Y| do(\XX))$, is defined by marginalizing all the variables other than $Y$. 

\paragraph{Causal-effect queries.} If we have access to the full causal model we can answer causal-effect queries from the full model 
using traditional probabilistic inference. However,
normally we have no access to the full 
model $\M$. Yet,
it was shown that assuming only a causal graph it is  
often possible to evaluate 
the effect of an intervention from the observational distribution $P(\V)$ only. This occurs if the answer is unique for any full model that is consistent with the graph and $P(\V)$ \cite{pearlbook}. More formally:
\begin{definition}[Identifiability, causal-effect]
 Given a causal diagram $\G$, the 
 {\bf causal-effect} query $P({\YY} \mid do({\XX}))$ is {\bf identifiable} if any two SCMs 
 having the same $\G$ that agree on the observational distribution $P(\V)$ 
 also agree on $P(\textbf{Y} \mid do(\textbf{X}))$. 
\end{definition}

\noindent Identifiability makes the causal-effect query well-posed.

\shrink{
\noindent Identifiability makes the causal-effect query well-posed:

\begin{definition}[Causal-effect query]
Given a causal diagram $\G = \langle \V \! \cup \U, E \rangle$, data samples from an observational distribution $P(\V)$, 
the causal effect query is to compute the distribution 
$P({\YY} \mid do({\XX=x}))$ whenever the query is identifiable. 
\end{definition}
}
\paragraph{Estimand-based approaches.}
The now-standard methodology for answering causal-effect queries is to break the task into two steps.
The first is the \emph{identifiability step}: given a causal diagram and a query $P(\YY \mid do(\XX))$, determine if the query is identifiable and if so, generate an \emph{estimand}, or an algebraic expression 
in terms of the observational distribution $P(\V)$, that answers the query.
A complete polynomial algorithm called ID has been developed for this task \cite{jinthesis,shpitser2006identification}.
The second step is \emph{estimation}: use samples from the observational distribution $P(\V)$ to estimate the value of the estimand expression. 

A number of approaches have been applied to the estimation in the second step.
A simple approach, called the \emph{plug-in estimator}, replaces each term in the estimand with its empirical probability value in the observed data.
More sophisticated approaches include 
the line of work by Jung et al.,
which apply ideas from data weighting and empirical risk minimization
\cite{DBLP:conf/aaai/Jung0B20,DBLP:conf/nips/Jung0B20} 
and double or debiased machine learning \cite{DBLP:conf/aaai/Jung0B21,DBLP:conf/icml/Jung0B21}.
Evaluating the estimand's value from a PAC perspective is analyzed in 
\cite{pmlr-v151-bhattacharyya22a}.

In practice, the estimand expression typically consists of multiple products or ratios of partially marginalized distributions.  We illustrate with an example:
\begin{example}\label{ex1}
Consider the model in Figure \ref{fig:chain_scm}. To evaluate the query $P(V_6 \mid do(V_0))$, the ID algorithm \cite{jinthesis,shpitser2006identification} gives the expression: %
\begin{multline} \label{eq:smallchainexp}
    P(V_6 \mid do(V_0))
    = \!\!\!\! \sum\limits_{V_1, V_2, V_3, V_4, V_5} \!\!\!\!\!\! P(V_5 | V_0, V_1, V_2, V_3, V_4)\\
    \times P (V_3 | V_0, V_1, V_2) \, P(V_1 | V_0) \sum\limits_{v_0} P(V_6 | v_0, V_1, V_2, V_3, V_4, V_5 ) \\
    \times P(V_4 | v_0, V_1, V_2, V_3) \,P(V_2 | v_0, V_1) \,P(v_0).   
\end{multline}
\end{example}
Unfortunately, in large models the expression elements quickly become unwieldy;
Example~\ref{ex1} involves estimating conditional distribution terms over the entire joint configuration space of the model. In terms of scalability, this presents several practical problems as model sizes increase.
The required distributions have exponentially many configurations, suggesting they may require a significant amount of data to estimate accurately.
Moreover, the required marginalizations are over high-dimensional spaces, potentially 
making them computationally intractable.
These issues make it difficult to apply 
statistically sophisticated or machine learning-based estimators 
\cite{DBLP:conf/aaai/Jung0B20,DBLP:conf/nips/Jung0B20,DBLP:conf/aaai/Jung0B21,DBLP:conf/icml/Jung0B21}
in such settings.

On the other hand, we can often maintain tractability by using the simple \emph{plug-in} estimator, in which each term is estimated only on the configurations seen in the observed data.
This reduces computation, since each term is non-zero on only a subset of configurations, and summations can also be performed over only the non-zero configurations.
However, this approach also limits the quality of our estimates -- our plug-in estimates may have high variance, especially in settings with very large probability tables,
and it is difficult to apply regularization or other variance-reduction techniques without destroying the sparsity property that makes it computationally feasible.

\paragraph{Treewidth.} The treewidth of a graph (also known as induced-width) characterizes how close is the graph to a tree structure. If the treewidth of a graph is $w$, the graph can be embedded in a tree (or a hypertree) of clusters whose sizes are bounded by $w$.
Exact PGM inference is exponential in the treewidth of their graphs. For more formal definition and related properties see \cite{DBLP:journals/ai/Dechter99,DBLP:journals/ai/KaskDLD05,DBLP:series/synthesis/2013Dechter}.

The above described issues on the estimand's computational and estimation challenges, motivate us to explore the effectiveness of a \emph{model completion} approach, in which we directly learn the Causal Bayesan Network model, including its latent variables, and then use the resulting learned model to evaluate the query.
A common technique for learning latent variable models is the EM algorithm.

\paragraph{Expectation Maximization (EM).}
EM is a well-studied scheme to learn Bayesian network parameters when 
the graph itself is known but there are some latent variables, for which data are missing \cite{dempster1977,koller}. \nocite{DarwicheBook09}
EM is an iterative maximum likelihood approach that alternates between  
an Expectation, or E-step, and a Maximization, or M-step;
see \cite{em} for details.
The E-step entails inference in the Bayesian network, which can be computationally hard in general; but when the graph has bounded \emph{treewidth}, the E-step can be performed using join tree or bucket elimination algorithms \cite{em,DBLP:series/synthesis/2013Dechter}. Otherwise, approximate algorithms such as belief propagation can be used \cite{Pearl89b,DBLP:journals/jair/MateescuKGD10}\nocite{conf/nips/YedidiaFW00}. 
While EM has also been applied to structure learning \cite{DBLP:journals/corr/abs-1301-7373},
in our setting the structure is assumed known via the provided causal diagram.

\paragraph{Other related work.}  
Motivated by similar model completion ideas, \cite{DBLP:journals/corr/abs-2202-02891,DBLP:conf/uai/ChenD22} 
first convert the causal diagram to a circuit, exploiting the functional form of SCM mechanisms to yield compact circuits, then learn the circuit parameters via EM, and answer the causal queries.  
However, their work is restricted to binary-valued variables and, to the best of our knowledge, no empirical evaluation is provided.
Another related line of work explores neural network approaches for causal inference \cite{DBLP:conf/nips/XiaLBB21,DBLP:journals/corr/abs-2210-00035}. In that work the authors focus on \emph{counterfactual} queries and propose learning a Neural Causal Model (NCM) whose functions are neural networks, a task which 
requires learning the mechanisms of the underlying SCM, which is
significantly more challenging and unnecessary for our task of answering causal-effect queries. Moreover, while the authors provide a theoretical analysis, empirical validation is only performed over very small synthetic networks.
EM has also been applied to causality in~\cite{DBLP:journals/corr/abs-1202-3763,DBLP:journals/corr/abs-2011-02912};
in the latter, the authors assume knowledge of the functional mechanisms in the SCM, with only the distributions of the discrete latent variables unknown.
In contrast, our method assumes knowledge of the graph only.

\section{Learning-Based Causal Inference}\label{method}

The approach we propose to answer the causal-effect query is to first learn a full CBN $\B = \langle \G, \mathcal{P} \rangle$ from the given causal diagram $\G = \langle \V \! \cup \U, E \rangle$ and from samples from the observational distribution $P(\V)$. Then, we can answer causal-effect queries from the full model, employing the appropriate truncation 
Eq.~\eqref{eq-intervene},
via probabilistic inference algorithms over the learned CBN. 
However, there could be many parameterizations $\mathcal{P}$ that are consistent with the same observational distribution $P(\V)$.
Luckily, as long as the query is identifiable, any of these alternative parameterizations $\mathcal{P}$ will generate the same answer:

\begin{theorem}

\label{prop1}
Assume a given SCM $\M = \langle \U, \V, \F, P(\U) \rangle$ having causal diagram $\G$ and observational distribution $P(\V)$.
Any CBN $\B = (\G, \mathcal{P})$ inducing the same observational distribution $P(\V)$ via Eq.~\eqref{eq-obs} will agree with $\M$ on any identifiable causal-effect query $P({\YY} \mid do({\XX}=\xx))$. 
\begin{proof} Follows directly from the definition of identifiability.
\end{proof}
\end{theorem}

\noindent

Consequently, our problem reduces to the well-studied task of estimating the parameters of a Bayesian network from the network structure when given data on a subset of latent variables is not available. A widely used algorithm for this purpose is Expectation-Maximization (EM), which aims to maximize the likelihood of the data. To apply EM, however, we also require knowledge of the latent variables' distributional forms.

Usefully, in the case of discrete visible variables $\V$, prior work has shown that any SCM can be transformed into an equivalent SCM in which the latent variables $\U$ take on discrete, finite domains \cite{DBLP:conf/icml/Zhang0B22} and it provides an upper bound on the required latent domain sizes.
We will therefore assume a discrete distribution for the $U_j$.

However, the upper bound in \cite{DBLP:conf/icml/Zhang0B22} is conservative, and often unnecessarily large. 
So, we decided to use the discrete latent domain sizes as a complexity control mechanism, to impose an additional degree of regularity on the probability distribution over the visible variables. 
For regularization, we treat the domain sizes, 
$\kk = \{k_{U_i}\}$, as hyperparameters and select their values by minimizing the model's BIC score, which 
penalizes models with larger domain sizes for their increased flexibility and potential over-fitting \cite{DarwicheBook09}:
\begin{equation}
 BIC_{\B,\D}= -2\cdot LL_{\B,\D} + p \cdot \log(|\D|)   
\end{equation}
where $\D$ are the data samples from $P(\V)$, $LL_{\B,\D}$ is the log likelihood of the CBN model $\B$ learned via EM, and $p$ is the number of free probability parameters, denoted $\btheta$, 
in the Bayesian network $\B$.

To optimize our CBN over both the domain sizes $\kk$ and probability parameters $\btheta$,
we propose a practical algorithm that searches greedily in the space of $\kk$ while optimizing $\btheta$ by EM. 
We prioritize searching the model space starting from smaller domain sizes since simpler models are preferable (i.e., Occam Razor). We adopt a simple strategy of keeping all latent domain sizes equal, i.e., set all $k_{U_i}=k$, and
gradually increase the value $k$.

Our learning strategy benefits from two sources of variance reduction compared to the simple plug-in estimates: first, from the graph structure, which imposes some regularity on $P(\V)$; and second, by preferring smaller latent domain sizes 
when possible, which encourages learning lower-rank distributions when supported by the data.

\paragraph{Model-completion rationale.}  Given any causal effect query $P({\YY} \mid do({\XX}=\xx))$ defined relative to causal graph $\G$ and given a full CBN $\B = \langle \G, \mathcal{P} \rangle$ learned from observational data, Theorem \ref{prop1} suggests that 
$P({\YY} \mid do({\XX}=\xx))$  can be estimated as $P({\YY})$ in the truncated learned CBN $\B_{\XX=\xx}$, i.e., 
\begin{equation}
P({\YY} \mid do({\XX}=\xx)) \approx
P_{\B}({\YY} \mid do({\XX}=\xx)) = P_{B_{\XX=\xx}} ({\YY}).
\label{eq3aa}
\end{equation}
where
$\B_{\XX=\xx} = \langle \G_{\overline{\XX}}, \mathcal{P}_{\XX=\xx} \rangle$ 
is the truncated CBN and 
$\G_{\overline{\XX}}$ is the graph obtained by deleting all incoming arrows to $\XX$ in $\G$.

\begin{algorithm}[t]
    \caption{EM4CI }
    \label{alg:em4ci}
    \Input{ A causal diagram $\G = \langle \U \! \cup \! \V, E \rangle$, $\U$  latent and $\V$ observable; samples $\D$ from $P(\V)$;\\ Identifiable Query $ Q= P({\YY} \mid do({\XX=\xx}))$.
     }
    \Output{Estimated $P({\YY} \mid do({\XX=\xx}))$}
    \hrulealg
    \vspace{6pt}
    \tcp{$k$= latent domain size, $BIC_{\B} \equiv BIC_{\B, \D}$ the $BIC$ score of \\ CBN $\B$ \& $\D$, $LL_{\B} \equiv LL_{\B, \D}$  the log-likelihood of $\B$ \& $\D$ }
    \DontPrintSemicolon
    \SetAlgoHangIndent{\widthof{Step 1:~~~~}}
    ~~1.\hspace{0.1cm}
Initialize:   $BIC_{\B} \leftarrow \infty $\\
    \vspace{2pt}
    ~~2.\hspace{0.1cm}
     {\bf for} $k = 2, ...,$ to upper bound, {\bf do}\\
     ~~3.~~~~Initialize: $LL_{\B^\prime} \leftarrow 0$\\
     ~~4.~~~~{\bf for} $i= 0\ldots 9$ {\bf do} $~~~~$ (optimize LL of EM)\\
     ~~5.~~~~~~ Initialize: $\theta_i $ to random parameters of $\G$\\     
  ~~6.~~~~~~ $(LL_{\B_{i}}, \B_i) \leftarrow \mathrm{EM}(\G, \D, k, \theta_i)$  \!\!\!\!\!\!\;  
  ~~7.~~~~~~ {\bf if} $LL_{\B_{i}} > LL_{\B^\prime}$ \\
  ~~8. ~~~~~~~~$LL_{\B^\prime} \leftarrow LL_{\B_{i}}, \B^\prime \leftarrow \B_{i}   $\\
     ~~9.~~~~{\bf end for}\\
 ~~10.~~~~ Calculate $BIC_{\B^\prime}$ from $LL_{\B^\prime}$\\
 ~~11.~~~~ {\bf if} $BIC_{\B^\prime} \leq BIC_{\B}$, \\
~~12. ~~~~~~~~~$\B \leftarrow \B^\prime$, $BIC_{\B} \leftarrow BIC_{\B^\prime}$ \\
~~13. ~~~~{\bf else},  break.\\
\vspace{2pt}
    ~~14. ~~{\bf endfor}\\
    ~~15:\hspace{0.1cm}
    $\B_{{\XX=\xx}}$ $\leftarrow$ truncated CBN from $\B$.\;
    \vspace{2pt}
    ~~16:\hspace{0.1cm}
    \Return  
     $P_{\B_{\XX=\xx}}({\YY})$\ computed by any PGM algorithm 
     \shrink{
         \Return ~~ $\leftarrow \mathrm{PGM\ inference}(\B_{\XX=\xx}, P({\YY}))$\;
         }
    \vspace{2pt}
   
\end{algorithm}

\paragraph{The iterative learning EM4CI algorithm (Algorithm \ref{alg:em4ci}).}
Since we use EM for learning we call our algorithm {\em EM for Causal Inference (EM4CI)}, described in Algorithm \ref{alg:em4ci}. Its input is a causal diagram $\G$,  samples $\D$ from $P(\V)$,  
and an identifiable query $Q=P({\YY} \mid do({\XX=\xx}))$. 
\textit{Steps 2-14} provide the iterative learning scheme. 
Given samples $\D$ from $P(\V)$, we use EM to learn a CBN $\B$ over the graph $\G$ with
latent domain size $k$. EM outputs the parameters for a network denoted $\B_i$, and its corresponding log-likelihood, $LL_{\B_i}$.
Since EM's performance is known to be sensitive to initialization, we perform ten runs of EM starting from randomly generated initial parameters, and retain the model, ${\B^\prime}$, with the highest likelihood, $LL_{\B^\prime}$,  as the candidate for the current latent domain size $k$ (\textit{steps 4-9}), with which we calculate the $BIC_{\B^\prime}$ score (\textit{step 10}). 
If the $BIC$ score decreases (\textit{steps 11-13}), we adopt the new candidate network as $\B$, and the process continues with increased $k$; otherwise the current $\B$ is selected as the final learned network. 
\textit{Step 15} generates the truncated CBN $\B_{\XX=\xx}$, and \textit{step 16} employs a standard {\em PGM} inference algorithm, like join-tree decomposition \cite{DarwicheBook09,DBLP:series/synthesis/2013Dechter}, to the CBN $\B_{\XX=\xx}$ to compute 
$P({\YY})$. 



\paragraph{Complexity of EM4CI.} 
The complexity of EM ({\em step 6}) depends on the sample size $|D|$, the variables' domain sizes, and the number of iterations needed for convergence which is hard to predict, but 
 complexity-wise provides only a constant factor and can be ignored.  \textit{Steps 2-14} repeat the EM algorithm, which again adds only a constant factor.
The most extensive computation in each iteration of EM is the Expectation step, which requires probabilistic inference and can be accomplished in time and memory exponential in the treewidth of the graph. Otherwise, approximation algorithms such as belief propagation may be used. In our experiments, exact inference was always possible. 
Thus, each iteration of EM takes $O(|D| \cdot n l^w)$, where $n= |V \cup U|$ is the number of variables,  
$l$ bounds the variable domain sizes, and $w$ is the treewidth. Thus for  $T$ iterations we have 
$O(T\cdot |D| \cdot n \cdot l^w)$. 
{\em Step 16} of probabilistic inference is $O(n \cdot l^w)$ with exact inference. In summary, the complexity of algorithm {\em EM4CI} is $O(T \cdot |D| \cdot n \cdot l^w)$. Thus, the algorithm is most effective for models with bounded treewidth. 
Note that the cost of the EM learning process can be amortized over multiple queries. In summary,

\begin{theorem}[complexity of EM4CI]
The complexity of algorithm {\em EM4CI} is $O(T \cdot m \cdot n \cdot l^w)$, where $m$ is the number of samples, $T$ is the number of iterations, $k$ bounds the domain size, $n$ is the total number of variables, and $w$ is the treewidth of the causal diagram.
\end{theorem}

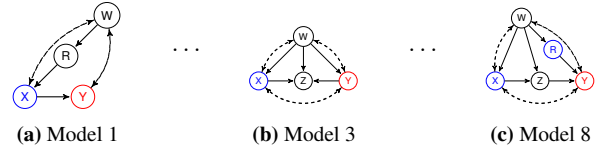
\begin{figure}[t]
     \centering
     \begin{subfigure}[b]{0.28\columnwidth}
           \centering
           \resizebox{.6\linewidth}{!}{\input{tikzfigures/model1.tikz}}  
           \caption{Model 1}
           \label{fig:model1}
      \end{subfigure}
      \hfill  
      \raisebox{1cm}{$\ldots$}
      \hfill
          \begin{subfigure}[b]{.28\columnwidth}
           \centering
           \resizebox{.6\linewidth}{!}{\input{tikzfigures/model3.tikz}}  
           \caption{Model 3}
           \label{fig:model3}
      \end{subfigure}
      \hfill  
      \raisebox{1cm}{$\ldots$}
      \hfill
          \begin{subfigure}[b]{.28\columnwidth}
           \centering
           \resizebox{.6\linewidth}{!}{\input{tikzfigures/model8.tikz}}  
           \caption{Model 8}
           \label{fig:model8}
      \end{subfigure}
      \\
      \vspace{5mm}
     \caption{A subset of the small causal diagrams for models used in our experiments. \\[2em] }
     \label{fig:allmodels}
 \end{figure}

\begingroup

\renewcommand{\arraystretch}{1.25}

\begin{table}[tb]
    \centering
     \Huge
    \caption{Estimand Expressions for Models 1-8 in Figure~\ref{fig:allmodels} \& Table \ref{tab:results}. }
    \resizebox{\columnwidth}{!}{
    \begin{tabular}{ c c}
    \toprule
    Model & Estimate of {$P(Y \mid do(X))$} \\
    \midrule
         1 & $\frac{\sum_W P(X,Y |R,W)P(W)}{\sum_W P(X |R,W)P(W)}$\\  
         2 & $\sum_R P(R|X_1) \sum_{x_1,Z} P(Y|R,x_1,X_2,Z)P(Z|R,x_1)P(x_1)$ \\
         3 & $P(Y) $\\
         4 & $\sum_{R,S} P(S | X_1,X_2,X_3,Z)P(R|X_1)\sum_{x_1,x_3,Z} P(Y|R,x_1,X_2,x_3,Z)P(x_3|x_1,X_2,Z)P(x_1,Z)$ \\
         5 & $\sum_{Z_1,Z_2,Z_3} P(Z_3|Z_2)P(Z_1|X,Z_2) \frac{\sum_{x}P(Y,Z_3|x,Z_1,Z_2)P(x,Z_2)}{\sum_{x}P(Z_3|x,Z_1,Z_2)P(x,Z_2)} P(Z_2)$ \\
          6 & $\sum_{Z_2} P(Y|X_1,X_2,Z_2)P(Z_2) $\\
         7 & $\sum_{Z_2,Z_3} P(Y|X_1,X_2,Z_2,Z_3)P(Z_2,Z_3) $\\
         8 & $\sum_{R,W,Z} P(Z|R,W,X)P(R|W)\sum_{x} P(Y|R,W,x,Z)P(x|R,W)P(W) $\\
    \bottomrule
        \end{tabular}}
    \label{tab:modelexp}
\end{table}
\endgroup

\section{Empirical Evaluation}\label{eval}
We perform an extensive empirical evaluation of EM4CI and compare
against estimand-based approaches such as the brute-force empirical \emph{plug-in} method and a state-of-the-art scheme called 
weighted empirical risk minimization 
(WERM) \cite{DBLP:conf/nips/Jung0B20}. All
experiments were run on a 2.66 GHz processor with 8 GB of memory.  All experimental code can be found \href{https://github.com/anniemeow/EM4CI}{here} \cite{github}.

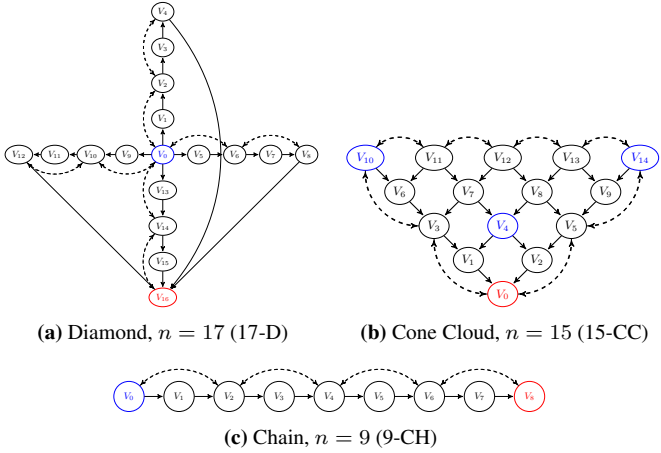
\begin{figure}[t] \centering
     \begin{subfigure}[b]{0.48\columnwidth} \centering
        \resizebox{\linewidth}{!}{\input{tikzfigures/diamond.tikz}}  
        \caption{Diamond, $n=17$ (17-D)\\}
           \label{fig:diamond}
      \end{subfigure}
      \hfill  
     \begin{subfigure}[b]{0.48\columnwidth} \centering
        \resizebox{\linewidth}{!}{\input{tikzfigures/cone_cloud.tikz}} 
        \caption{Cone Cloud, $n=15$ (15-CC)\\}
           \label{fig:cone_cloud}
      \end{subfigure}\\
      \vspace{15pt}
         \begin{subfigure}[b]{0.66\columnwidth} \centering
        \resizebox{\linewidth}{!}{\input{tikzfigures/chain_9.tikz}} 
        \caption{Chain, $n=9$ (9-CH)\\}
           \label{fig:chain_9}
      \end{subfigure}
      \vspace{5mm}
     \caption{Larger synthetic models.\\}
     \vspace{10mm}
\end{figure}

\subsection{Benchmarks.}
The inputs to a causal inference algorithm are a triplet: 1) a causal diagram (of an underlying SCM), 2) data from the model's observational distribution, and 3) a query $Q = P(\YY \mid do(\XX))$. 
We use two sources for our benchmarks: synthetically generated scalable model classes, and real domains from the academic literature of various fields with causal interpretations \cite{smileman}.

\paragraph{Synthetic networks.}
We generate each triplet benchmark instance by first choosing a causal diagram and a query. Then, picking domain sizes of observed and latent variables, we generate the conditional probability tables (CPTs), one per variable and its parents, to yield the full CBN.
From this full model we generate samples 
by forward sampling \cite{DarwicheBook09} 
and then discard the values of the latent variables, yielding samples from the observed distribution. We compute the exact answer to the target query on the full CBN via an exact algorithm (e.g., join-tree  \cite{DarwicheBook09,DBLP:series/synthesis/2013Dechter}).

The causal diagrams 
were selected in two ways. First, we use a set of 8 small diagrams from the causality literature  \cite{pearlbook,DBLP:conf/aaai/Jung0B21}\nocite{DBLP:conf/nips/Jung0B20}; see Figure \ref{fig:allmodels} and the Supplemental \cite{supplement}.
Second, we examine three scalable classes of graphs 
whose treewidths vary but can be controlled.
These are: chain networks  (treewidth 3; Figure \ref{fig:chain_9}), diamond networks (treewidth 5; Figure \ref{fig:diamond}), and cone-cloud networks (treewidth  $O(\sqrt{n})$,  
Figure \ref{fig:cone_cloud}), abbreviated CH, D, CC respectively.

In the small diagrams (Figure \ref{fig:allmodels}), we set the domain size of the observed variables to $d=2$, and of the latent variables to $k=10$.
For the chain, diamond, and cone-cloud models, we use $(d,k)=(4,10)$.
The parameters of each CPT were generated by sampling 
from a Dirichlet distribution \cite{DarwicheBook09}
with parameters
$\alpha = [ \alpha_1, \ldots, \alpha_n ]$, where 
$\alpha_i\in [0,1)$ selected uniformly at random. The aim is
to generate conditional probabilities near the edges of the simplex (i.e., far from uniform).

\paragraph{Use-case benchmarks.} 
We also experimented with four networks created for real-world domains.
The ``Alarm'' network was developed for on-line monitoring of patients in intensive care units \cite{alarm}; 
the ``A'' network, from the UAI literature, is synthetic but known to be daunting to exact algorithms \cite{a-net};
the ``Barley'' network was built for a decision support system for growing barley without pesticides \cite{barley}; 
and the ``Win95'' model is an expert system for printer troubleshooting in Windows 95\nocite{win95}.
Since no variables in these networks are specified as latent, 
we designated source vertices with at least two children as latent confounding variables. As in the synthetic networks case, we generated data by forward sampling, subsequently discarding the latent variables' values.

\paragraph{Queries.} We hand selected 
multiple identifiable queries per model, aiming for non-trivial ones and prioritizing those that appear to be complex. We provide the queries and their corresponding estimands in the Supplemental \cite{supplement}.

\begin{table}[t]
   \caption{ Algorithms and Packages}
    \centering
      \resizebox{\columnwidth}{!}{%
    \begin{tabular}{ | c c c c |}
    \toprule
    Algorithm & Use & Software Package & Programming Language\\
    \midrule
    ID Algorithm & Plug-In \& EM4CI & {\em causaleffect} & R\\
    Join-tree Decomposition & EM4CI inference & SMILE & C++\\
    EM Algorithm & EM4CI learning & SMILE & C++\\
    Sparse Plug-In Evaluation & Plug-In &  & Python\\
    WERM & & & R \\
    \bottomrule
    \end{tabular}}   
    \vspace{1em}
     \label{tab:tools}
\end{table}

\subsection{Algorithms and performance measures.}

\paragraph{EM4CI.} 
In our EM4CI algorithm we used the EM implementation of SMILE: Structural Modeling, Inference, and Learning Engine package \cite{smileman}, written in C++. 
In SMILE, inference is carried out via join trees \cite{Pearl89b,em}\nocite{DBLP:series/synthesis/2013Dechter}, and used both in the E step of the EM algorithm \cite{em} and for answering queries over the learned model.
 
\paragraph{Estimand-based algorithms: Plug-In and WERM.}
We used the ID algorithm in the causal effect package \cite{causaleffect} to compute the estimands.
Subsequently, the plug-in method evaluates the estimand from the observational data, by plugging in the empirical conditional probabilities. Our plug-in implementation uses sparse table representations in Python, to ensure that the representation size of each estimated probability term is linear in the data size, rather than exponential in the number of variables.
We also compare against a state-of-the-art scheme called WERM  \cite{DBLP:conf/nips/Jung0B20} using the author-provided implementation in R.
A summary of the algorithms and implementations is given in Table \ref{tab:tools}.

\paragraph{Measures of performance.} 
We report the results of EM4CI along the two phases of the algorithm; the {\em learning phase} (steps 2-14) done via EM and the {\em inference phase} (steps 15-16) done by the join tree algorithm. For the learning process, we report the selected latent domain sizes and the total time at termination. 
For the inference phase we report the
{\em mean absolute deviation (mad)}  
between the true answer 
and the estimated answer. 
The measure $mad$ for a query $P({Y} \mid do({\XX}))$ is computed by averaging the absolute error over all 
instantiations of the intervened and queried variables, 
$P({Y=y}\mid do({\XX}=\xx))$ for $\xx,y \in \D({\XX})\times\D(Y)$.
We also report inference time for each query.
We incremented $k$ by steps of two ($k = 2, 4,6, 8, 10 \ldots)$ in the EM4CI algorithm.

\begin{table}[t]
   \caption{ Results of EM4CI \& Plug-In on  $P(Y| do(\textbf{X}))$  $(d,k) = (2,10)$}
    \resizebox{\columnwidth}{!}{%
    \begin{tabular}{ c | c c c | c c | c c c |  c c  }
    \toprule 
      &  \multicolumn{5}{c|}{100 Samples} & \multicolumn{5}{c}{1,000 Samples} \\
    \cmidrule{2-11}
      &  \multicolumn{3}{c}{EM4CI} &  \multicolumn{2}{c|}{Plug-In} & \multicolumn{3}{c}{EM4CI} &  \multicolumn{2}{c}{Plug-In} \\
    Model& $k_{lrn}$ & mad & time(s)
    &  mad  & time(s) & $k_{lrn}$ & mad & time(s) & mad & times(s) \\
        \midrule
        1 & 2  & 0.0037 & 0.48 & 0.0104 & 1.9 & 2  & 0.0032 & 3.1 & 0.0025 & 2.3 \\     
        2 & 2  & 0.1832 & 1.86 & 0.1436 & 2.3 & 2  & 0.0490 & 8.4 & 0.0867 & 2.0 \\
        3 & 2  & 0.1288 & 0.93 & 0.0569 & 1.1 & 2  & 0.0040 & 3.6 & 0.0039 & 0.7 \\
        4 & 2  & 0.1819 & 1.82 & 0.1469 & 2.3 & 2  & 0.1438 & 12.0 & 0.0704 & 2.1 \\
        5 & 2  & 0.4910 & 1.65 & 0.5000 & 2.0  &2  & 0.0044 & 17.3 & 0.0058 & 2.2 \\
        6 & 2 & 0.2663 & 0.30 & 0.3930 & 2.0 & 2  & 0.1263 & 0.5 & 0.1319 & 2.1 \\
        7 & 2  & 0.2520 & 0.78 & 0.2509 & 1.9 & 2  & 0.0891 & 7.1 & 0.0238 & 2.0 \\
        8 & 2  & 0.1372 & 0.63 & 0.1579 & 2.0 & 2  & 0.2340 & 4.7 & 0.1303 & 1.9 \\
    \bottomrule
    \end{tabular}}
    \label{tab:results}
\end{table}

\begin{table}[t]
    \caption{Results of absolute error on Query $P(Y=1 | do (\textbf{X}=1))$ on Models 1, 8, \& 3' 
    by WERM \& EM4CI. 
    $k_{lrn}$ is the learned domain sizes of latent variables.}
    \resizebox{\columnwidth}{!}{
     \begin{tabular}{ c | c c | c  c c | c c | c c c  }
    \toprule 
      &  \multicolumn{5}{c|}{1,000 Samples} & \multicolumn{5}{c}{10,000 Samples} \\
    \cmidrule{2-11}
      &  \multicolumn{2}{c|}{WERM} &  \multicolumn{3}{c|}{EM4CI} & \multicolumn{2}{c|}{WERM} &  \multicolumn{3}{c}{EM4CI} \\
    Model& error& time(s) & error & time(s) & $k_{lrn}$ 
    & error & time(s) & error & time(s) & $k_{lrn}$ \\

        \midrule
         1 & 0.0071 & 18.7 &  0.0059 &8.8 &2 & {0.0031} & 32.6 & 0.0046 & 63.5 &2  \\ 
         8 & 0.1082 & 25.8 & 0.1566 & 7.6 &2 & 0.11 & 47.7&  0.0001& 81.4 & 2  \\
        3' & 0.027 & 27.2 &  0.0004&  3.5&   2  & 0.001 & 44.1 &0.0009 & 53.1 &2 \\

    \bottomrule
    \end{tabular}}
    \vspace{1em}
    \label{tab:werm}
\end{table}

\begin{table}[t]
   \caption{ Results for EM4CI and Plug-In on  $P(Y| do(\textbf{X}))$  $(d,k) = (4,10)$}
    \vspace{4mm}
\begin{subtable}{\columnwidth}
   \caption{1,000 samples}
    \resizebox{\columnwidth}{!}{%
    \begin{tabular}{ c c| c c c c | c  c  }
    \toprule 
      &  &\multicolumn{4}{c|}{EM4CI} &  \multicolumn{2}{c}{Plug-In} \\
       Model &Query& $k_{lrn}$  & mad & learn-time(s) & inf-time(s) &  mad  & time(s) \\
         \midrule
   5-CH &$P(V_4 | do(V_0))$ & 4 & 0.0902 & 3.5 & 0.0001 & 0.1509 & 2.3 \\
    9-CH& $P(V_8 | do(V_0))$& 4  & 0.1204 & 11.5 & 0.0002 & 0.1516 & 2.4 \\
    25-CH& $P(V_{24} | do(V_0))$ & 2  & 0.0070 & 77.7 & 0.0003 & 0.0959 & 6.1 \\
    49-CH&$P(V_{48} | do(V_0))$ & 4  & 0.0005 & 161.2 & 0.0007 & 0.0319 & 17.8 \\
    99-CH& $P(V_{98} | do(V_0))$ & 6  & 0.0093 & 413.4 & 0.0023 & 0.0611 & 88.1 \\
    9-D& $P(V_8 | do(V_0))$  & 2  & 0.0719 & 24.6 & 0.0002 & 0.1832 & 3.4 \\
    17-D& $P(V_{16} | do(V_0))$  & 6  & 0.0542 & 202.3 & 0.0006 & 0.0700 & 4.5 \\
    65-D& $P(V_{64} | do(V_0))$  & 4  & 0.0074 & 432.4 & 0.0012 & 0.1716& 232.5 \\
    6-CC& $P(V_0 | do(V_5))$ & 4 & 0.0088 & 23.5 & 0.0001 & 0.0156 & 2.3 \\
    15-CC& $P(V_0 | do(V_{14}))$ & 4 & 0.0147 & 60.8 & 0.0001 & 0.0659 & 4.5 \\
    45-CC& $P(V_0 | do(V_{14}, V_{36}, V_{44}))$  & 6  & 0.0097 & 199.2 & 2.7429 & 0.1509 & 18.6 \\
 
    \bottomrule
    \end{tabular}}
    \label{tab:compare1000}
        \vspace{4mm}

\end{subtable}
\begin{subtable}{\columnwidth}
  \caption{10,000 samples}
    \resizebox{\columnwidth}{!}{%
    \begin{tabular}{ c c | c c  c c | c c   }
    \toprule 
      &  & \multicolumn{4}{c|}{EM4CI} &  \multicolumn{2}{c}{Plug-In} \\
    Model & Query & $k_{lrn}$  & mad & learn-time(s) & inf-time(s) &  mad  & time(s) \\
        \midrule
   5-CH & $P(V_4 | do(V_0))$ &4& 0.0508 & 17.3 & 0.0001 & 0.0537 & 2.5 \\
    9-CH & $P(V_8 | do(V_0))$& 4 & 0.0236 & 150.0 & 0.0002 & 0.1074 & 3.1 \\
    25-CH & $P(V_{24} | do(V_0))$ &  6& 0.0068 & 697.1 & 0.0005 & 0.0714 & 26.4 \\
    49-CH & $P(V_{48} | do(V_0))$ & 10 & 0.0017 & 2412.6 & 0.0036 & 0.0160 & 133.7 \\
    99-CH & $P(V_{98} | do(V_0))$ & 6 & 0.0028 & 3887.9 & 0.0022 & 0.0433 & 850.6 \\
    9-D & $P(V_8 | do(V_0))$  & 4 & 0.0611 & 390.7 & 0.0002 & 0.1481 & 3.0 \\
    17-D & $P(V_{16} | do(V_0))$& 6  & 0.0360 & 1849.6 & 0.0007 & 0.0582 & 8.4 \\
    65-D & $P(V_{64} | do(V_0))$  &4 & 0.0022 & 4787.2 & 0.0013 & 0.1376 & 2258.5 \\
    6-CC & $P(V_0 | do(V_5))$ & 6  & 0.0138 & 116.9 & 0.0003 & 0.0136 & 2.7 \\
    15-CC & $P(V_0 | do(V_{14}))$ &4  & 0.0022 & 489.5 & 0.0043 & 0.0321 & 10.9 \\
    45-CC &$P(V_0 | do(V_{14}, V_{36}, V_{44}))$ & 6 & 0.0026 & 1833.7 & 2.757 & 0.1561 & 105.8 \\
  
    \bottomrule
    \end{tabular}}
        
    \label{tab:compare10000}
\end{subtable}
\label{tab:compare_all}
\end{table}

\subsection{Results}

\paragraph{Results on small synthetic models.}
Results on models 1-8 (Figure \ref{fig:allmodels}) are presented in Table $\ref{tab:results}$. Since these models are quite small 
we report the total time (learning plus inference) for EM4CI. 
We compare against the {\em Plug-In} estimand method (see estimand expressions in Table \ref{tab:modelexp}), which is guaranteed to converge to the exact answer. Therefore, we expect {\em Plug-In} to produce fairly accurate results on these small models if given enough samples.
We observe that the accuracy of both methods are similar at both 100 and 1,000 samples, with EM4CI being more accurate on some cases, and {\em Plug-In} on others. EM4CI had better time performance for 100 samples, 
but for 1,000 samples the {\em Plug-In} due to the learning time of EM4CI.

\paragraph{WERM comparison.}
The results for Models 1, 8, and 3' \footnote{model 3' is the same as Model 3 in Figure \ref{fig:allmodels}, but with edge $Y \rightarrow Z$ 
reversed  
to match the hard-coded model in WERM.}
comparing WERM \cite{DBLP:conf/nips/Jung0B20} to EM4CI are given in Table~\ref{tab:werm}. We use domain size $d=2$, 
and 1,000 and 10,000 samples.
Again, EM4CI produced more accurate results in several instances, though the disparities are smaller than with the Plug-In method. 
EM4CI was faster than WERM with 1,000 samples but slower with 10,000 samples. 
Unfortunately, the available
code for WERM is specific to these models, so we were unable to compare against it more generally.

\begin{figure*}[t]
        \resizebox{\linewidth}{!}{\input{tikzfigures/3graphs.tikz}}  
    
         \caption{Comparing the accuracy trends of EM4CI and Plug-In as a function of the model size. 
     While both methods improve with more samples (solid to dashed lines),
     the error ($mad$) of EM4CI is smaller, even when compared to Plug-In with more samples.
     }
     \vspace{5mm}
     \label{fig:madgraphs}
\end{figure*}
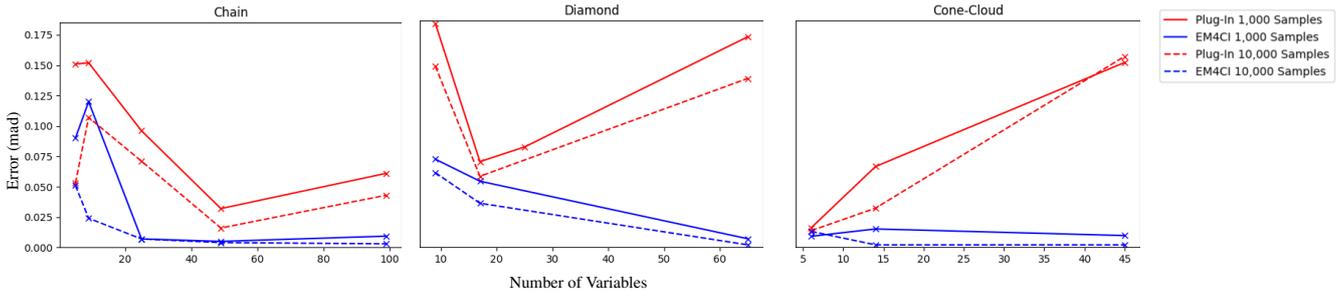

\begingroup
\begin{table}[t]
       \centering
           \captionsetup{justification=centering}
       \caption{  {EM4CI \& Plug-In on the 45 Cone Cloud} \\ $|\V| =45$; $ ~|\U|=16$; $d=4 ; k=10$; treewidth $\approx 11$\\}
       \vspace{4mm}
\begin{subtable}{\columnwidth}
    \caption{Plug-In}
    \scriptsize    
      \resizebox{\columnwidth}{!}{%
    \begin{tabular}{  c | c c | c c   }
    \toprule
    &\multicolumn{2}{c|}{\textit{1,000 Samples}} & \multicolumn{2}{c}{\textit{10,000 Samples}} \\       \cmidrule{2-5}
      Query& mad &  time(s) & mad &  time(s)\\
      \midrule
    $P(V_0|do(V_{14}, V_{36}, V_{44}))$ & 0.1510 & 18.6 & 0.1561 & 105.8 \\
$P(V_0|do(V_{12}, V_{29}, V_{34}))$ & 0.1744 & 19.0 & 0.1043 & 101.0 \\
$P(V_0|do(V_{10}, V_{14}, V_{40}))$ & 0.1906 & 18.4 & 0.1201 & 131.4 \\
$P(V_4|do(V_{15}, V_{20}, V_{40}))$ & 0.2080 & 26.2 & 0.1282 & 121.6 \\

    \bottomrule
    \end{tabular}}
    \label{tab:45conecloud_plug_in}
    \vspace{4mm}

\end{subtable}\\
\begin{subtable}{\columnwidth}   
   \caption{ EM4CI}
    \centering
      \resizebox{\columnwidth}{!}{%
    \begin{tabular}{  c | c c |c c  }
    \toprule
    &\multicolumn{2}{c|}{\textit{1,000 Samples}} & \multicolumn{2}{c}{\textit{10,000 Samples}} \\
    \cmidrule{2-5}
   \textbf{Learning:}&  time(s) $=199$ & $k_{lrn}=6$ &    time(s)$ = 1833$  & $k_{lrn}=6$\\[.5em]
   \midrule
   \textbf{Inference:}&&&& \\
      Query& mad &  time(s) & mad &  time(s)\\
      \midrule
    $P(V_0|do(V_{14}, V_{36}, V_{44}))$ & 0.0097 & 2.77  & 0.0026 & 2.76 \\
$P(V_0|do(V_{12}, V_{29}, V_{34}))$ & 0.0118 & 18.99 & 0.0024 & 18.99 \\
$P(V_0|do(V_{10}, V_{14}, V_{40}))$ & 0.0097 & 0.06 & 0.0029 & 0.06 \\
$P(V_4|do(V_{15}, V_{20}, V_{40}))$ & 0.0082 & 0.02 & 0.0013 & 0.02 \\

    \bottomrule
    \end{tabular}}
    \label{tab:45conecloud_em4ci}

\end{subtable}
\label{tab:45cone_cloud}
\end{table}
\endgroup

\paragraph{Results on large synthetic models.} 
In Tables \ref{tab:compare_all} and \ref{tab:45cone_cloud} we show results on larger models of chains, diamonds, and cone-cloud graphs. The first two tables compare EM4CI to the plug-in method for a single query over all the models. Specifically, Table \ref{tab:compare1000} presents time and accuracy results for 1,000 samples. We see that EM4CI was consistently more accurate, and in many cases significantly better (e.g., in 45-cone-cloud).  Table \ref{tab:compare10000} shows the same trend for 10,000 samples. 
The superior accuracy of learning for model completion over {\em Plug-In} is clearly visible in Figure \ref{fig:madgraphs}, which shows the accuracy trends for each  class as a function of model sizes, for both 1,000 and 10,000 samples. We see that generally, EM4CI using only 1,000 samples is even more accurate than {\em Plug-In} with 10,000 samples.

Again, we expect that this improvement is due to the variance reduction of the estimation process.  By its nature, model completion exploits more information from the causal graph than is apparent in the estimand expression alone, implicitly capturing the known conditional independence structure in its estimates of $P(\V)$.  We can also include simple and easy-to-impose complexity control, in the form of the latent domain sizes, to further reduce variance. In contrast, it is difficult to impose meaningful regularity or variance reduction on the plug-in estimates of $P(\V)$ without creating significant computational burdens.
Taking it all together, our results suggest that whenever the causal graph's treewidth is bounded and the estimand expression has large scope functions, we should prefer using model completion. 
Other settings may require more study to determine the best approach.

Focusing on time, 
we see that the time of EM4CI is significantly larger than {\em Plug-In}, with EM learning being the most time consuming part. Interestingly, while time grows with model size for both schemes, inference time 
remains efficient, likely due to the low treewidth of some of the models (e.g., the chains and the diamonds). In the 45-cone case, inference is impacted more by model size, since its treewidth increases with the square root of the number of variables.  
Generally, for a single query, we find that the {\em Plug-In} is superior time-wise, and its time increases at a slower pace with model size. 
Finally, in both methods, time increases with sample size as expected.

\begin{table}[b]
  
   \caption{ Results on EM4CI \& Plug-In on Real  Networks}
   \vspace{4mm}
   \begin{subtable}{\columnwidth}
   \caption{1,000 Samples}
    \resizebox{\columnwidth}{!}{%
    \begin{tabular}{ c c| c c c c | c  c  }
    \toprule 
      &  &\multicolumn{4}{c}{EM4CI} &  \multicolumn{2}{c}{Plug-In} \\
       Model &Query& $k_{lrn}$  & mad & learn-time(s) & inf-time(s) &  mad  & time(s) \\
         \midrule
 
    A& $P(V_{51}|do(V_{10}))$& 4 & 0.0139 &71 & 0.0012 & 0.0584 & 8\\
    Alarm&$P(HRBP|Shunt))$& 2 & 0.0076 & 15.7 & 0.0002 & 0.0190 & 2.9 \\
    Barley& $P(protein|do(expYield))$& 14 & 0.0066 & 199   & 0.9038 & 0.0687 & 9.9\\
    Win95& $P(Output|do(NnPsGrphc))$& 2 & 0.0113  & 109.1 &  0.0002 &0.2674 & 1.5\\

    \bottomrule
    \end{tabular}}
    \vspace{4mm}
\end{subtable}
\begin{subtable}{\columnwidth}
  \caption{10,000 samples}
    \resizebox{\columnwidth}{!}{%
    \begin{tabular}{ c c | c c  c c | c c   }
    \toprule 
      &  & \multicolumn{4}{c}{EM4CI} &  \multicolumn{2}{c}{Plug-In} \\
    Model & Query & $k_{lrn}$   & mad & learn-time(s) & inf-time(s) &  mad  & time(s) \\
        \midrule
 
    A & $P(V_{51}|do(V_{10}))$& 4 & 0.0083 & 540.6 & 0.0012  & 0.0114 & 55.7\\
    Alarm & $P(HRBP|Shunt))$ & 4 & 0.0043 & 181.3 & 0.0002 & 0.0075 & 5.7\\
    Barley& $P(protein|do(expYield))$ &10 &0.0031 & 819.5 & 0.8882  & 0.0655 &63.4\\
    Win95 &  $P(Output|do(NnPsGrphc))$ & 4 & 0.0008 & 1080.8 & 0.0002 &0.2894 &2.0\\
    \bottomrule
    \end{tabular}}
        
\end{subtable}
\label{tab:compare_real}
\vspace{2mm}
\end{table}

\paragraph{Results on real-world data sets.}    
Table \ref{tab:compare_real} compares EM4CI against {\em Plug-In} on a single query 
for all 4 networks. We observe the same pattern of performance as in the synthetic networks: EM4CI provides far more accurate results on all real-world models but with some time cost, attributed to learning.
Evaluation for multiple queries 
are given in  Table \ref{tab:A-results} for the \textbf{ $\mathbf{A}$ network}.
We find that the same latent domain size ($k_{lrn} = 4$) is selected for both sample sizes. As before, learning time grows with sample size and accuracy results are excellent and are improving with increased sample sizes. 
Note, however, that increased sample size
does not impact inference time. Results for {\bf Alarm}, {\bf Win95}, and {\bf Barley} networks are similar and can be found in the supplemental \cite{supplement}.

\paragraph{Answering multiple queries.}
Tables \ref{tab:45conecloud_em4ci} and
\ref{tab:A-em4ci} highlight how learning time 
can be amortized effectively over multiple queries.  The table presents EM4CI's learning and inference time separately on the multiple queries.
In contrast, the {\em Plug-In} method requires estimation of each new query from scratch, even if on the same model; see Table \ref{tab:45conecloud_plug_in} and \ref{tab:A_plugin}. 
Thus, for multiple queries, EM4CI may take far less time per query, while providing superior quality estimates.
(See additional results on multiple queries in the Supplemental \cite{supplement}.) 

 \paragraph{Summary.}
 Our experiments illustrate 
 that model-completion by learning implemented in EM4CI yields highly accurate estimates of causal effect queries  on a variety of benchmarks, including both synthetic and real-world networks of varying sizes. Moreover EM4CI shows clear superiority to the Plug-In estimands approach. In terms of time efficiency however, EM4CI was consistently slower due to learning time overhead. Yet, when answering multiple queries is desired  the time overhead can be amortized over multiple queries.

 \begingroup
\begin{table}[b]
\vspace{2mm}
\centering
    \captionsetup{justification=centering}
\caption{Plug-In \& EM4CI results on the \textbf{A Network}\hspace{1mm}\\
$|\V| =46 ;~|\U|=8;~d=2;~k=2$, treewidth $\approx 16$}
\vspace{4mm}
\begin{subtable}{\columnwidth}
    \caption{Plug-In}
    \scriptsize 
      \resizebox{\columnwidth}{!}{%
    \begin{tabular}{  c c c |c c   }
    \toprule
    &\multicolumn{2}{c|}{\textit{1,000 Samples}} & \multicolumn{2}{c}{\textit{10,000 Samples}} \\       \midrule
      Query& mad &  time (s) & mad &  time (s)\\
      \midrule
  $P(V_{51} | \text{do}(V_{10}))$ & 0.0584 & 8.0 &0.0114 & 55.7 \\
$P(V_{51} | \text{do}(V_{14}))$ & 0.0319& 8.3&0.0056 & 51.3 \\
$P(V_{51} | \text{do} (V_{41}))$ & 0.0255& 13.9& 0.0092& 48.3\\
$P(V_{51}  | \text{do} ({V_{45} }))$ & 0.0496& 9.8&0.0206 & 49.1  \\

    \bottomrule
    \end{tabular}}
    \label{tab:A_plugin}
    \vspace{4mm}

\end{subtable}\\
\begin{subtable}{\linewidth}
              \captionsetup{justification=centering}

   \caption{ EM4CI}
    \centering
      \resizebox{\columnwidth}{!}{%
    \begin{tabular}{  c c c |c c  }
    \toprule
    &\multicolumn{2}{c|}{\textit{1,000 Samples}} & \multicolumn{2}{c}{\textit{10,000 Samples}} \\
    \midrule
      \textbf{Learning} & $time (s)=71$ & $k_{lrn}=4$ &    $time (s)=541$  & $k_{lrn}=4$\\
       \midrule
        \textbf{Inference:}&&&& \\
Query& mad &  time (s) & mad &  time (s)\\
      \midrule
   $P(V_{51} | \text{do}(V_{10}))$ & 0.0139 & 0.0012 & 0.0083 & 0.0012 \\
$P(V_{51} | \text{do}(V_{14}))$ & 0.0143 & 0.0047 & 0.0086 & 0.0046 \\
$P(V_{51} | \text{do} (V_{41}))$ & 0.0147 & 0.0042 & 0.0079 & 0.0041 \\
$P(V_{51}  | \text{do} ({V_{45} }))$ & 0.0140 & 0.0031 & 0.0082 & 0.0030 \\
      
    \bottomrule
    \end{tabular}}
    \label{tab:A-em4ci}
    \vspace{4mm}

\end{subtable}\\%

\label{tab:A-results}
\end{table}
\endgroup

\section{Conclusion} \label{conclusion}
In causal inference, the estimand-based approach has become standard: generating a potentially complex expression in terms of the observable distribution, and then estimating the required probabilities quantities from the data and evaluating the resulting expression.
While mathematically correct, this approach tends to ignore the difficulties inherent in estimating the complex, conditional probabilities required, and the computationally challenge in evaluating the resulting expression. These difficulties become increasingly apparent as model size increases.

An alternative path, relatively less explored, is to leverage the causal model structure more directly via learning.  By performing model completion -- i.e., learning a Causal Belief Network, including its latent confounding variables, from the observed data -- we exploit additional information about the co-dependence structure in the distribution, and can more easily apply complexity control and variance reduction strategies to the parameter estimation process. Then, once an approximate full model is available, traditional computationally efficient PGM algorithms can be applied to answer the query, either exactly or approximately.

Our algorithm EM4CI uses the well-known EM algorithm to learn the model and its latent variable distributions, then uses tree-decomposition algorithms to answer the queries.

We carried out an extensive empirical evaluation, the first of its scale in the causal community, over a collection incorporating both synthetic networks and real world distributions. We evaluated and compared EM4CI's performance to the Plug-In estimand approach, in terms of both accuracy and time.

Our results appear conclusive: we show clearly that model completion using EM4CI yields consistently superior results compared to the estimand Plug-In. This benefit does come with a cost in time from the learning phase, which grows more quickly with network size and number of samples compared to the Plug-In approach. However,  learning time can be amortized over multiple queries on the same model, making a collection of queries significantly more efficient.

Our EM4CI approach relies in part on the efficiency of inference; when the treewidth of the graph is bounded, both learning and inference using EM4CI are likely to be effective.
Additionally, since the structure of the graph is better exploited by model completion, we reduce the variance of our estimators, resulting in better performance for a given dataset size.
Although in some cases, the estimand approach may generate a simple and easy-to-estimate expression, in larger models with many latent confounders the expression is often complex.
In such settings, the estimand expression loses structural information, forcing us to estimate high-dimensional probabilities and making it difficult to apply variance reduction strategies without creating a computationally difficult evaluation problem.
Overall this suggests that model completion should be considered a competitive strategy for causal estimation.

\section{Acknowledgments}
Thank you to the reviewers for their valuable comments and
suggestions. This work was supported in part by NSF grants 
IIS-2008516 and CNS-2321786.


\clearpage

\clearpage
\bibliography{ecai24}
\newpage
\onecolumn
\appendix
\section{Supplementary Material}
\renewcommand{\baselinestretch}{1.5} 
\subsection{Models}
Here we show the additional small models used in our experiments;
see Figures 2 \& 3 in the main text.
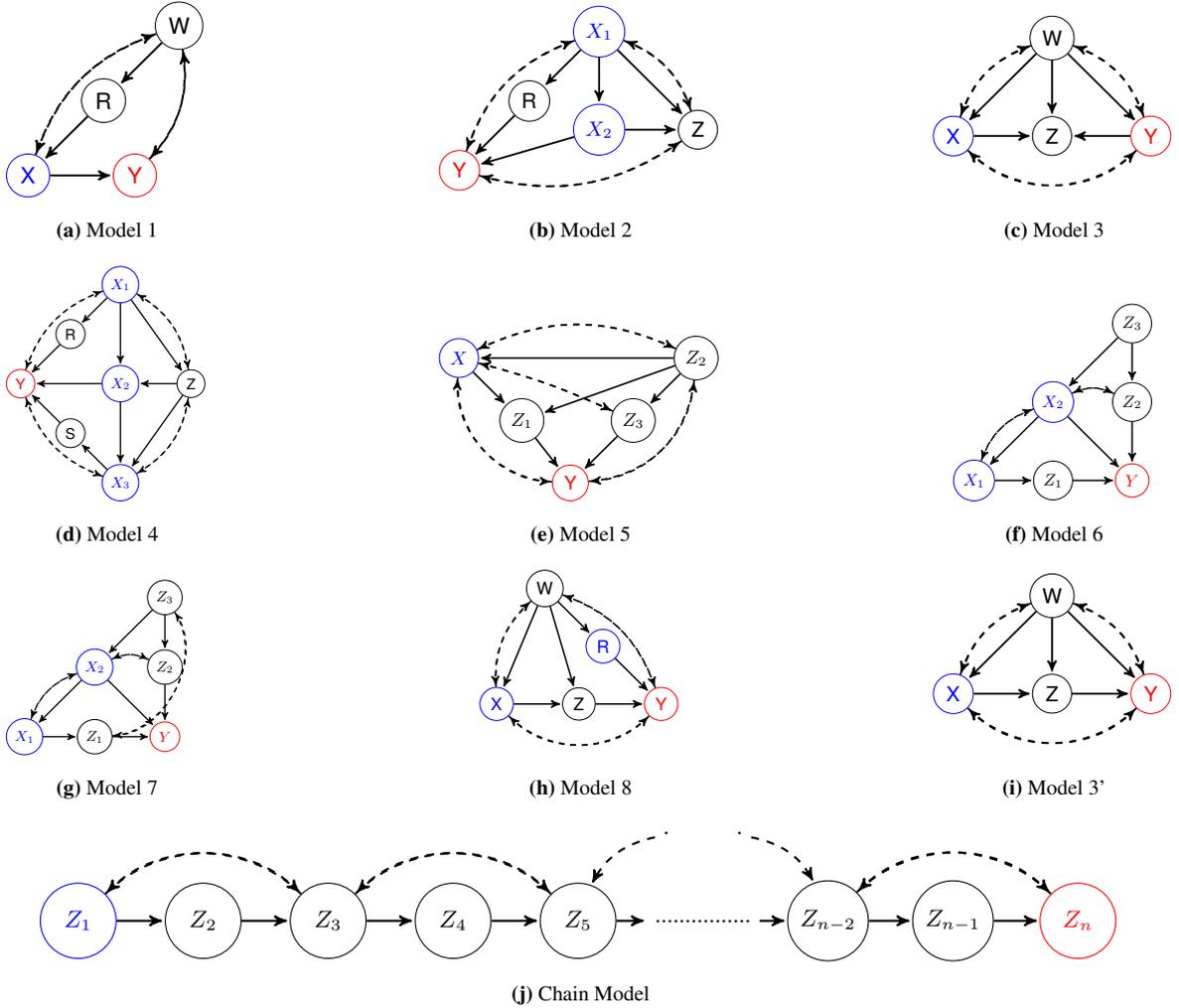
\begin{figure}[h]
     \centering
    
     \begin{subfigure}[b]{0.30\textwidth}
           \centering
           \resizebox{.5\linewidth}{!}{\input{tikzfigures/model1.tikz}}  
           \caption{Model 1}
           \label{fig:model1}
      \end{subfigure}
      \hfill
     \begin{subfigure}[b]{0.30\textwidth}
           \centering
           \resizebox{.7\linewidth}{!}{\input{tikzfigures/model2.tikz}}  
           \caption{Model 2}
           \label{fig:model2}
      \end{subfigure}
      \hfill
       \begin{subfigure}[b]{0.30\textwidth}
           \centering
           \resizebox{.6\linewidth}{!}{\input{tikzfigures/model3.tikz}}             \caption{Model 3}
          \label{fig:model3}
      \end{subfigure}
      \\\vspace{5mm}
     \begin{subfigure}[b]{0.30\textwidth}
           \centering
           \resizebox{.5\linewidth}{!}{\input{tikzfigures/model4.tikz}}  
           \caption{Model 4}
           \label{fig:model4}
      \end{subfigure}
      \hfill
          \begin{subfigure}[b]{0.30\textwidth}
           \centering
           \resizebox{.7\linewidth}{!}{\input{tikzfigures/model5.tikz}}  
           \caption{Model 5}
           \label{fig:model5}
      \end{subfigure}
      \hfill
       \begin{subfigure}[b]{0.30\textwidth}
           \centering
           \resizebox{.5\linewidth}{!}{\input{tikzfigures/model6.tikz}}  
           \caption{Model 6}
           \label{fig:model6}
      \end{subfigure}
      \\\vspace{5mm}
     \begin{subfigure}[b]{0.3\textwidth}
           \centering
           \resizebox{.5\linewidth}{!}{\input{tikzfigures/model7.tikz}}  
           \caption{Model 7}
           \label{fig:model7}
      \end{subfigure}
      \hfill
          \begin{subfigure}[b]{.3\textwidth}
           \centering
           \resizebox{.5\linewidth}{!}{\input{tikzfigures/model8.tikz}}  
           \caption{Model 8}
           \label{fig:model8}
      \end{subfigure}
      \hfill
          \begin{subfigure}[b]{.3\textwidth}
           \centering
           \resizebox{.6\linewidth}{!}{\input{tikzfigures/model9.tikz}}  
           \caption{Model 3'}
           \label{fig:model9}
      \end{subfigure}
\\
\vspace{5mm}
      \begin{subfigure}[b]{\textwidth}
           \centering
           \resizebox{.8\linewidth}{!}{\input{tikzfigures/chain.tikz}}  
           \caption{Chain Model}
           \vspace{5mm}

           \label{fig:chainmodel}
      \end{subfigure}
      \\
    
     \caption{Causal diagrams for models used for experiments. Blue variables are intervened on and red variables are the outcome variables corresponding to the query $P(Y \mid do (X))$. \\
     }
     \label{fig:allmodels}
     \vspace{10mm}
\end{figure}

\subsection{Additional Results}

We provide several additional tables of results to complement those presented in the main text on larger synthetic and real networks. In Table \ref{tab:99chain} and \ref{tab:65diamond}, we see the results of EM4CI on the {\bf 99-chain} and {\bf 65-diamond} on multiple queries. We see the same trend discussed in the paper, where learning time is longer than inference time, and grows with sample size. Again, the learned domain sizes are the same for both sample sizes, and close to the true domain size of the latent variables. We also see that inference is very fast and therefore we can amortize the learning time over multiple queries if desired. Finally, we see that EM4CI is very accurate.

In Table \ref{tab:Alarm-results} we see results on the Plug-In method and EM4CI on the {\bf Alarm network}. Again, we see that EM4CI is more accurate than the Plug-In, but the learning time is longer. However, for multiple queries, EM4CI may take less time per query, while providing superior quality estimates.

Lastly in Table \ref{tab:all_real} the results for EM4CI on the {\bf Barley  network} and the {\bf Win95} are displayed in  the context of multiple queries, illustrating a similar pattern.  For the {\bf Barley network} (Table \ref{tab:Barley-results}), the learned domain sizes are large ($k_{lrn} =14, 10$) for each sample size, and accordingly learn time is also large for both settings. However, inference time remains very low. 
\begingroup
\begin{table}[t]
\caption{\large{EM4CI on large synthetic models}}
\vspace{5mm}
\hspace{15mm}
\begin{subtable}{.35\linewidth}
        \captionsetup{justification=centering}
        \caption{{\bf 99 chain}:\\ $|\V| =99; ~|\U|=49 ;d=4 ; k=10$. treewidth=2}
    \centering
      \resizebox{\columnwidth}{!}{%
    \begin{tabular}{  c c c |c c   }
    \toprule
    &\multicolumn{2}{c|}{\textit{1,000 Samples}} & \multicolumn{2}{c}{\textit{10,000 Samples}} \\
    \midrule
     \textbf{Learning}&  time(s)  & $k_{lrn}$ &    time(s)  & $k_{lrn}$\\
    \midrule
      & 413.4 & 6  &   3887.9& 6 \\
      \bottomrule
       \textbf{Inference}&  &   & &\\
       \midrule
      Query& mad &  time(s) & mad &  time(s)\\
      \midrule
    $P(V_{98} | do(V_{0}))$ & 0.0093 & 0.0023& 0.0028 & 0.0022\\
    $P(V_{49} | do(V_{0}))$ & 0.0113 & 0.0011 & 0.0041 & 0.0011 \\
     $P(V_{98} | do(V_{49}))$ & 0.0093 & 0.0011 &  0.0028 & 0.0011 \\
     $P(V_{98} | do(V_{90}))$ & 0.0152 & 0.0004 & 0.0063 & 0.0004 \\

    \bottomrule
    \end{tabular}}
        \label{tab:99chain}
\end{subtable}\hfill 
\begin{subtable}{.35\linewidth}
        \captionsetup{justification=centering}
        \caption{{\bf 65 diamond:}\\ $|\V| =65; ~|\U|=32; d =4 ; k=10$. treewidth=5}
    \centering
      \resizebox{\columnwidth}{!}{%
    \begin{tabular}{  c c c |c c   }
    \toprule
    &\multicolumn{2}{c|}{\textit{1,000 Samples}} & \multicolumn{2}{c}{\textit{10,000 Samples}} \\
    \midrule
    \textbf{Learning}&  time(s) & $k_{lrn}$&   time(s)  & $k_{lrn}$\\
    \midrule
        &432.3& 4  &   4787.2& 4\\
      \bottomrule
       \textbf{Inference}&  &  &  &\\
       \midrule
      Query& mad &  time(s) & mad &  time(s)\\
      \midrule
    $P(V_{64} | do(V_{0}))$ & 0.0074 & 0.0012 &  0.0022 & 0.0013 \\
    $P(V_{32} | do(V_{16}))$ & 0.0193 & 0.0004 &  0.0046 & 0.0005 \\
    $P(V_{16} | do(V_{0}))$ & 0.0283 & 0.0004 &  0.0150 & 0.0005 \\
    $P(V_{48} | do(V_{32}))$ & 0.0086 & 0.0004 &  0.0093 & 0.0004 \\

    \bottomrule
    \end{tabular}}
        \label{tab:65diamond}

\end{subtable}
\hspace{15mm}

\label{tab:all_synthetic}
\end{table}
\endgroup
\begingroup
\begin{table}[b]
\captionsetup{justification=centering}
\caption{\large Plug-In \& EM4CI results on the \textbf{Alarm Network}\hspace{1mm}\\
$|\V| =32 ;~|\U|=5;~2\leq d \leq4;~ 2 \leq k\leq 3$. treewidth$\approx 3 $}
\vspace{5mm}
\begin{subtable}{.5\linewidth}
    \caption{Plug-In}
      \resizebox{\columnwidth}{!}{%
    \begin{tabular}{  c c c |c c   }
    \toprule
    &\multicolumn{2}{c|}{\textit{1,000 Samples}} & \multicolumn{2}{c}{\textit{10,000 Samples}} \\       \midrule
      Query& mad &  time(s) & mad &  time(s)\\
      \midrule
   $P(HRBP | \text{do(Shunt)})$ & 0.0190 & 2.9&	0.0075 & 5.7 \\
$P(HRBP | \text{do(VentAlv)})$ & 0.0433 &	3.5	&0.0212 & 5.7\\
$P(HR | \text{do(VentAlv)})$ & 0.04706 &4 & 0.0184 & 	5.53\\
$P(HR | \text{do(Shunt)})$ & 0.0229	&3.0	&0.00296	&6.4\\

    \bottomrule
    \end{tabular}}
    \label{tab:Alarm_plugin}
    \vspace{15mm}

\end{subtable}
\begin{subtable}{.5\linewidth}
    \centering
    \captionsetup{justification=centering}
    \vspace{5mm}
   \caption{ EM4CI}
      \resizebox{\columnwidth}{!}{%
    \begin{tabular}{  c c c |c c  }
    \toprule
    &\multicolumn{2}{c|}{\textit{1,000 Samples}} & \multicolumn{2}{c}{\textit{10,000 Samples}} \\
    \midrule
      \textbf{Learning}&  time(s) & $k_{lrn}$ &  time(s)  & $k_{lrn}$\\
    \midrule
     & 16 & 2 & 181 & 4 \\
      \bottomrule
       \textbf{Inference}&   & & &\\
       \midrule
      Query& mad &  time(s) & mad &  time(s)\\
      \midrule
      $P(HRBP | \text{do(Shunt)})$ & 0.0076 & 0.0002 & 0.0043 & 0.0002 \\
$P(HRBP | \text{do(VentAlv)})$ & 0.0146 & 0.0003 & 0.0033 & 0.0003 \\
$P(HR | \text{do(VentAlv)})$ & 0.0106 & 0.0002 & 0.0020 & 0.0003 \\
$P(HR | \text{do(Shunt)})$ & 0.0101 & 0.0002 & 0.0027 & 0.0002 \\

    \bottomrule
    \end{tabular}}
     \label{tab:Alarm-em4ci}

\end{subtable}\\%

\label{tab:Alarm-results}
\end{table}

\begingroup
\begin{table}[b]
\caption{\large EM4CI results on the Real Networks\\}
\vspace{10mm}
\begin{subtable}{.5\linewidth}
    \captionsetup{justification=centering}

   \caption{ "{\bf Barley}":\\$|\V| =42; ~|\U|=6;~2 \leq d \leq 67 ;2 \leq k \leq  9 $. treewidth$\approx 7$ }
    \centering
      \resizebox{\columnwidth}{!}{%
    \begin{tabular}{  c c c |c c   }
    \toprule
    &\multicolumn{2}{c|}{\textit{1,000 Samples}} & \multicolumn{2}{c}{\textit{10,000 Samples}} \\
    \midrule
     \textbf{Learning}&  time(s) & $k_{lrn}$&  time(s)  & $k_{lrn}$\\
    \midrule
     & 199 & 14 &  820 & 10 \\
      \bottomrule
       \textbf{Inference}&  &  & &\\
       \midrule
      Query& mad &  time(s) & mad &  time(s)\\
      \midrule
        $P(\text{Protein} | \text{do(expYield)})$ & 0.0066 & 0.9038 & 0.0031 & 0.8882 \\
    $P(\text{FieldCap} | \text{do(protein)})$ & 0.0108 & 0.0004 & 0.0032 & 0.0004 \\
    $P(\text{expYield} | \text{do(precipitation)})$ & 0.0284 & 0.0002 & 0.0064 & 0.0002 \\
    $P(\text{weight} | \text{do(precipitation)})$ & 0.0107 & 0.0002 & 0.0023 & 0.0002 \\ 
      
    \bottomrule
    \end{tabular}}
        \label{tab:Barley-results}
\end{subtable}\hfill
\begin{subtable}{.48\linewidth}
              \captionsetup{justification=centering}

   \caption{"{\bf Win95}":\\$|\V| =59; ~|\U|=17;~d=2 ;k=2$ treewidth$\approx 8$ }
    \centering
      \resizebox{\columnwidth}{!}{%
    \begin{tabular}{  c c c |c c  }
    \toprule
    &\multicolumn{2}{c|}{\textit{1,000 Samples}} & \multicolumn{2}{c}{\textit{10,000 Samples}} \\
    \midrule
     \textbf{Learning}&  time(s) & $k_{lrn}$ &    time(s)  & $k_{lrn}$\\
    \midrule
    & 109 & 2 & 1081 & 4 \\
      \bottomrule
       \textbf{Inference}&  &  & & \\
       \midrule
      Query& mad &  time(s) & mad &  time(s)\\
      \midrule
      $P(\text{Ouput} | \text{do(NnPsGrphic)})$ & 0.0113 & 0.0002 & 0.0008 & 0.0002 \\
$P(\text{PrintData} | \text{do(localOK)})$ & 0.0768 & 0.0002 & 0.0049 & 0.0002 \\
$P(\text{PCtoPRT} | \text{do(netOK)})$ & 0.0167 & 0.0001 & 0.0141 & 0.0002 \\
$P(\text{PrintData} | \text{do(lnetOK)})$ & 0.0116 & 0.0002 & 0.0016 & 0.0002 \\
    \bottomrule
    \end{tabular}}
        \label{tab:Win95-results}
\end{subtable}
\label{tab:all_real}
\end{table}
\endgroup
\clearpage

\section{Estimand Expressions}
We also provide the estimand expressions for the queries presented in the paper. In some cases these expressions are impractically large to include directly, in which case we provide a table listing the number of variables in the largest summation and function (probability term). All expressions were derived using the Causal Effect package \cite{causaleffect}.

\subsection{Diamond Model}
First we present the expression on the diamond model when $|\V|=17$. This expression is quite large, so for $|\V|=65$ we summarize the estimand's properties in a table.

\begin{multline}
    P (V_{16} | do(V_{0})) =\\ \sum\limits_{V_1, V_2, V_3, V_4, V_5, V_6, V_7, V_8, V_9, V_{10}, V_{11}, V_{12}, V_{13}, V_{14,}, V_{15}} P ( V_{15},| V_{0}, V_{13}, V_{14}) \times\\\frac{\sum\limits_{V_0', V_2, V_6, V_{10}} numerator }{\sum\limits_{V_0', V_2, V_6, V_{10}}denominator}\times P(V_{13} | V_0) P ( V_{11} | V_0, V_9, V_{10} ) P(V_9 | V_0 ) \times \\
    P(V_7|V_0, V_5, V_6) P(V_5 | V_0) P(V_3 | V_0, V_1, V_2) P (V_1| V_0) \times \\
    \sum\limits_{V_0'} (V_{12} | V_0', V_9, V_{10}, V_{11} ) (V_{10} | V_0', V_9 ) P(V_0') \sum\limits_{V_0'} (V_{8} | V_0', V_5, V_{6}, V_{7} ) (V_{6} | V_0', V_5 ) P(V_0')  \\
    \times \sum\limits_{V_0'} (V_{4} | V_0', V_1, V_{2}, V_{3} ) (V_{2} | V_0', V_1 ) P(V_0') \hspace{50pt}
\end{multline}
\begin{multline}
    numerator =\\ P(V_{16}| V_0', V_1, V_2, V_3, V_4, V_5, V_6, V_7, V_8, V_9, V_{10}, V_{11}, V_{12}, V_{13}, V_{14,}, V_{15})\times \\
    P(V_{14}, V_4| V_0', V_1, V_2, V_3, V_5, V_6, V_7, V_8, V_9, V_{10}, V_{11}, V_{12}, V_{13}) \times\\
    P(V_{2}| V_0', V_1,  V_5, V_6, V_7, V_8, V_9, V_{10}, V_{11}, V_{12}, V_{13}) P(V_{8}| V_0',V_5, V_6, V_7, V_9, V_{10}, V_{11}, V_{12}, V_{13})  \\
   \times  P(V_{6}| V_0', V_5, V_9, V_{10}, V_{11}, V_{12}, V_{13}) 
    P(V_{12}| V_0',V_9, V_{10}, V_{11}, V_{13}) P(V_{10}| V_0', V_9, V_{13}) P(V_0') 
\end{multline}
\begin{multline}
    denominator =\\ P(V_{4}| V_0', V_1, V_2, V_3, V_5, V_6, V_7, V_8, V_9, V_{10}, V_{11}, V_{12}, V_{13}) \times \\
    P (V_2 |  V_0', V_1, V_5, V_6, V_7, V_8, V_9, V_{10}, V_{11}, V_{12}, V_{13}) P ( V_8 | V_0', V_5, V_6, V_7,  V_9, V_{10}, V_{11}, V_{12}, V_{13})  \\
 \times  P ( V_6 | V_0', V_5,  V_9, V_{10}, V_{11}, V_{12}, V_{13}) P ( V_{12} | V_0', V_9, V_{10}, V_{11}, V_{13}) P (V_{10} | V_0', V_9, V_{13} ) P(V_0') 
\end{multline}

\begin{table}[h]
    \centering

    \resizebox{.45\textwidth}{!}{
     \begin{tabular}{ c  c c  }
    \toprule 
    Query& size of  function & size of sum\\
        \midrule
    $P(V_{64}|do(V0))$ &65 & 63\\
    $P(V_{32}|do(V0))$ & 17& 16\\
    $P(V_{16}|do(V0))$ &17 &16\\
    $P(V_8|do(V0))$ & 17 &16\\

    \bottomrule\\
    \end{tabular}}
     \caption{Size of largest function \& sum in queries for Diamond model when $|\V|=65$}
    \label{}
\end{table}
\clearpage
\subsection{Cone-cloud Model}
Here we give the expression for a query the cone-cloud model when $|\V|=15$. Again, a table is provided summarizing the estimand for $|\V|=45$.
\begin{multline}
    P(V_0 | V_{14}, V_{10}, V_4 ) = \\
    \sum\limits_{V_1, V_2, V_3, V_5, V_6, V_7, V_8, V_9, V_{11}, V_{12}, V_{13}} P ( V_2 |  V_4, V_5, V_7, V_8, V_9, V_{11}, V_{12}, V_{13}, V_{14})\times \\
    P(V_9|V_{13}, V_{14}) P(V_8|V_{12},V_{13}) P(V_1| V_3, V_4, V_6, V_7, V_8, V_{10}, V_{11}, V_{12}, V_{13} )\times \\
    P(V_7| V_{11}, V_{12}) P(V_6 | V_{10},V_{11} ) P(V_{11}, V_{12}, V_{13}) \times \\
    \sum\limits_{V_{10}', V_{11}, V_{12}, V_{13}, V_{14}'} P(V_0 | V_1, V_2, V_3, V_4, V_5, V_6, V_7, V_8, V_9, V_{10}', V_{11}, V_{12}, V_{13}, V_{14}') \times \\
    P(V_5 | V_1, V_3, V_4, V_6, V_7, V_8, V_9, V_{10}', V_{11}, V_{12}, V_{13}, V_{14}') \times \\P(V_{14}' | V_1, V_3, V_4, V_6, V_7, V_8, V_{10}', V_{11},V_{12}, V_{13}) \times\\
    P(V_3, V_{13} | V_{6}, V_{7}, V_{10}', V_{12}, V_{13})  P(V_{10}' | V_{7}, V_{11}, V_{12}) P(V_{11}, V_{12}) \hspace{10pt}
\end{multline}

\begin{table}[h]
    \centering
    \resizebox{.45\textwidth}{!}{
     \begin{tabular}{ c  c c  }
    \toprule 
    Query& size of  function & size of sum\\
        \midrule
    $P(V_{0}|do(V_4,V_{36}, V_{44}))$ &45 & 41\\
    $P(V_{0}|do(V_{15},V_{20},V_{38},V_{42}))$ & 45& 34\\
    $P(V_{0}|do(V_4, V_{21}, V_{27}))$ &45 &37\\
    $P(V_0|do(V_{10}, V_{14}, V_{40}))$ & 45 &33\\

    \bottomrule\\
    \end{tabular}}
    \caption{Size of largest function \& sum in Queries for cone-cloud model when $|\V|=45$}
    \label{}
\end{table}
\clearpage
\subsection{A Model}
For illustration, we show one expression for the A model; the remainder are summarized via a table. 
\begin{multline}
    P(N_{51} | do(N_{45}) ) = \\
    \sum_{\substack{N_3, N_9, N_{10},N_{11}, N_{12}, N_{13}, N_{14},N_{15}, N_{17}, N_{18},N_{22},N_{23}, N_{24}, N_{25},N_{26}, \\ N_{27}, N_{28}, N_{29},N_{32},N_{34}, N_{35}, N_{36},N_{37},N_{39}, N_{41}, N_{43}, N_{47}, N_{49}, N_{50} }} P (N_{50} | N_3, N_9, N_{10},\ldots \\ N_{11}, N_{12}, N_{13}, N_{14}, N_{15}, N_{17}, N_{18}, N_{22}, N_{23}, N_{24}, N_{26}, N_{27}, N_{28}, N_{29}, N_{34}, N_{36},\ldots \\ N_{37}, N_{41}, N_{43}, N_{45}, N_{47}, N_{49}) \times\\ P(N_{47} | N_3, N_{11}, N_{13}, N_{14}, N_{17}, N_{18}, N_{22}, N_{23}, N_{24}, N_{26}, N_{27}, N_{36}, N_{43}, N_{45} ) \times \\
     P(N_{43}) P (N_{49} | N_{9}, N_{13}, N_{14}, N_{17}, N_{22}, N_{41}) P(N_{34} | N_{10}, N_{12}, N_{14}, N_{17}, N_{18}, N_{23}, N_{29}) \times \\
     P(N_{18})P(N_{32}| N_{10}, N_{11}, N_{12}, N_{14}, N_{15}, N_{17}, N_{25}) P(N_{26} | N_{14} ) P(N_{25} | N_{11}, N_{14}, N_{17} )\times \\
     P(N_{37} | N_{3}, N_{10}, N_{12}, N_{13}, N_{14}, N_{17}, N_{18}, N_{22}, N_{23}, N_{24}, N_{28}, N_{29}, N_{34}, N_{36} )  \times \\
     P(N_{22}| N_{3}, N_{10}, N_{12}, N_{13}, N_{14}, N_{17}) P ( N_{17} | N_{14}) \times \\
     P(N_{35} | N_{10}, N_{12})P(N_{27}|N_{11})P(N_{12}|N_{10})P(N_9)\\
     \sum_{V_{45}'} P(N_{51} | N_{3},N_9, N_{10}, N_{11}, N_{12}, N_{13}, N_{14},N_{15},  N_{17}, N_{18},N_{22},N_{23},N_{24}, N_{25}, \ldots \\N_{26},N_{27}, N_{28}, N_{29}, N_{32}, N_{34},N_{35},N_{36},N_{37}, N_{39},N_{41}, N_{43}, N_{45}', N_{47}, N_{49}, N_{50})   \times   \\
     P(N_{41}| N_{3},N_9, N_{10}, N_{11}, N_{12}, N_{13}, N_{14},N_{15},  N_{17}, N_{18},N_{22},N_{23},N_{24}, N_{25}, N_{26}\ldots \\
     N_{27}, N_{28}, N_{29}, N_{32}, N_{34},N_{35},N_{36},N_{37}, N_{39}, N_{45}') \times \\
     P(N_{45}'| N_{3},N_9, N_{10}, N_{11}, N_{12}, N_{13}, N_{14},N_{15},  N_{17}, N_{18},N_{22},N_{23},N_{24}, N_{25}, N_{26} \ldots \\
      N_{27}, N_{28}, N_{29}, N_{32},N_{34},N_{35},N_{36}, N_{39}) \times \\
      P(N_{23}, N_{29}| N_{3},N_9, N_{10}, N_{11}, N_{12}, N_{13}, N_{14},N_{15}, N_{17}, N_{18},N_{22},N_{24}, N_{25}, \ldots \\
      N_{26},N_{27}, N_{28}, N_{32}, N_{35}, N_{36}, N_{39}) \times \\
      P(N_{24}, N_{28}, N_{36}, N_{39} | N_{3},N_9, N_{10}, N_{11}, N_{12}, N_{13}, N_{14},N_{15},  N_{17} ,N_{22},N_{25}, \ldots \\
      N_{26}, N_{27}, N_{32}, N_{35}) \times \\
      P(N_{13}, N_{14} | N_{3},N_9, N_{10}, N_{11}, N_{12}, N_{15}, N_{27}, N_{35}) P(N_{15}| N_{3},N_9, N_{10}, N_{11},N_{12}, N_{27}) \times \\
      P(N_{11}| N_{3},N_9, N_{10},N_{12}) P(N_{10}| N_{3}, N_{9})P(N_3) \hspace{10pt}
\end{multline}
\begin{table}[h]
    \centering
    \resizebox{.45\textwidth}{!}{
     \begin{tabular}{ c  c c  }
    \toprule 
    Query& size of  function & size of sum\\
        \midrule
    $P(N_{51} | do(N_{10}) ) $ & 31 &29 \\
    $P(N_{51} | do(N_{14}) ) $ & 31 & 29\\
    $P(N_{51} | do(N_{41}) ) $ & 31& 29\\
    $P(N_{51} | do(N_{45}) ) $ & 31 &29\\

    \bottomrule\\
    \end{tabular}}
    \caption{    Size of largest function \& sum in queries for A-network where $|\V|=54$}
\end{table}
\clearpage
\subsection{Alarm Network}
The Alarm network has the following estimand expressions for the four evaluated queries:
$$P(HRBP|do(Shunt))= \sum_{ArtCO2}P(HRBP|Shunt, ArtCO2)P(ArtCO2)$$
$$P(HRBP|do(VentAlv))= \sum_{SaO2}P(HRBP|VentAlv, SaO2)P(SaO2)$$
$$P(HR|do(VentAlv))= \sum_{SaO2}P(HR|VentAlv, SaO2)P(SaO2)$$
$$ P(HR|Shunt) = \sum_{ArtCO2}P(HR|Shunt, ArtCO2)P(ArtCO2)$$
\subsection{Barley Network}
The Barley network has the following estimand expressions for the four evaluated queries:
\begin{multline}
    P(Protein|do(expYield))= \sum_{n4protein}P(n4protein |protein) \times \\
    \sum_{expYield'}P(Protein|expYield', n4protein)  \times P(expYield')
\end{multline}
\begin{multline}
    P(Protein|do(FieldCap))= \\
    \sum_{AllwedFertN}P(protein |FieldCap, AllowedFertN) \times P(AllowedFertN)
\end{multline}
$$P(yield|do(Precipitation))=P(Yield)$$
$$P(weight|do(Precipitation))= P(weight)$$
\subsection{Win95pts Network}
The Win95pts network has the following estimand expressions for the four evaluated queries:
$$P(Problem|do(NonPSGrphic))=P(Problem)$$
\begin{multline}
    P(Printdata|do(LocalOK))= \\
    \!\!\!\!\!\!\!\sum_{\substack{DOSlocalOK, DOSnetOK,\\GDIOutputOK ,netOK}} \!\!\!\!\!\!\!\!\!\!\!\!\!\!\!\!P(printData |localOK,DOSlocalOK,DOSnetOK,GDIOutputOK, netOk) \\\times P(DOSlocalOK,DOSnetOK,GDIOutputOK, netOk)
\end{multline}
\begin{multline}
    P(PCtoPRT|do(netOk))= \\
    \!\!\!\!\!\!\!\sum_{\substack{DOSlocalOK, DOSnetOK,\\GDIOutputOK ,localOK}} \!\!\!\!\!\!\!\!\!\!\!\!\!\!\!\!P(PCtoPRT |netOK,DOSlocalOK,DOSnetOK,GDIOutputOK, localOk) \\\times P(DOSlocalOK,DOSnetOK,GDIOutputOK, localOk)
\end{multline}
\begin{multline}
    P(printData|do(netOk))= \\
    \!\!\!\!\!\!\!\sum_{\substack{DOSlocalOK, DOSnetOK,\\GDIOutputOK ,localOK}} \!\!\!\!\!\!\!\!\!\!\!\!\!\!\!\!P(printData |netOK,DOSlocalOK,DOSnetOK,GDIOutputOK,localOk) \\\times P(DOSlocalOK,DOSnetOK,GDIOutputOK, localOk)
\end{multline}

\clearpage

\section{Code}
\textbf{To run EM4CI:}\\
All code can be found here \cite{github}
In order to run the code for this paper, a license to use BayesFusion software package SMILE is required.
EM4CI is written in C++. There is one source code file for the learning phase and one for inference. They are named: \begin{verbatim}
    learn_main.cpp   inf.cpp 
\end{verbatim}
In order to compile the source files, the command is:
\begin{verbatim}
    g++ -O3 learn_main.cpp -o learn.out -I./smile 
    -L./smile -lsmile
    g++ -O3 inf.cpp -o inf.out -I./smile 
    -L./smile -lsmile
\end{verbatim}
Which will produce the executables:
\begin{verbatim}
    learn.out         inf.out
\end{verbatim}
The {\bf learn.out } file expects command line arguments of the model file, the em-model file with a domain specified for the unobserved variables, a data csv file containing samples on the observed variables, and the number of samples. An example run is:
\begin{verbatim}
    ./learn.out models_xdsl/ex1_TD2_10.xdsl 
    models_xdsl/em_ex1_TD2_10_ED2_0.xdsl 
    data/100/ex1_TD2_10.csv  100
\end{verbatim}
The {\bf inf.out } file expects command line arguments of the model file, the learned model file that will be used to perform inference on, the query variable, $Y$, in $P(Y|do({\bf X}))$, the do variables(s) ${\bf X}$, and the number of samples used in the learning phase. An example run is:
\begin{verbatim}
    ./inf.out models_xdsl/ex1_TD2_10.xdsl 
    learned_models/100/em_ex1_TD2_10_ED2_0.xdsl 
    Y X 100
\end{verbatim}
The resulting log-likelihood, BIC score, time for learning, time for inference, and {\em mean absolute deviation (mad)}  are output to csv files, LL.csv, BIC.csv, timesLearn.csv, timesInf.csv, and err.csv, respectively. These will all be output to a folder named after the model, and containing a different subfolder per sample size and assumed domain size in the learning phase. For example, if running for assumed domain size 2 and sample size 100, the output will be in folder:
\begin{verbatim}
    ex1_TD2_10/100/em_ex1_TD2_10_ED2 
\end{verbatim}
The learned model files will be in the folder:
\begin{verbatim}
    learned_models/ex1_TD2_10/100 
\end{verbatim}
The bash script {\bf em4ci\_wrapper.sh} was used to automate the learning process. This script will automatically iterate through increasing latent domain sizes, while running the EM algorithm 10 times for each latent domain size. It will stop when the BIC score stops decreasing, and will output the minimum BIC score with the latent domain size of the final learned model.
To run this script you can pass in the model name and number of samples. For example:
\begin{verbatim}
    ./em4ci_wrapper.sh ex1_TD2_10 100
\end{verbatim}

The wrapper assumes all model files are contained in a folder named { \bf models\_xdsl} and data files are contained in a folder called {\bf data} with subfolders corresponding to the number of samples, like {\bf data/100}. 

The learned models files are of the form {\bf em\_ex1\_TD2\_10\_ED2\_0.xdsl } where the number after ED corresponds to the assumed domain size of the latent variables, and the last number corresponds to one of the runs $\{0, \ldots, 9\}$ that produced that model. You can perform inference on any learned model you like, but the {\bf em4ci\_wrapper.sh} outputs the run that correspond to the highest likelihood models with minimum BIC score, so we suggest using those.

All model files are in XDSL format, for more information see the Bayefusion Documentation
\cite{smileman}

\end{document}

%% file: tikzfigures/chain_7.tikz
\begin{tikzpicture}[->,>=stealth',shorten >=1pt,auto,node distance=1.65cm,
  main node/.style={circle,draw,minimum size=1cm,font=\sffamily\small}]
        
            \node[main node, color=blue] (1) { $V_0$};
            \node[main node] (2)[right of =1] { $V_1$};
            \node[main node] (3)[right of =2] { $V_2$};
            \node[main node] (4)[right of=3]{$V_3$};
            \node[main node] (5)[right of=4]{$V_4$};         
            \node[main node] (6)[right of=5]{$V_5$};
            \node[main node, color=blue] (7)[right of=6]{$V_6$};

                \draw[thick] (1) edge  (2);
               \draw[thick] (2) edge  (3);
                \draw[thick] (3) edge  (4);
               \draw[thick] (4) edge  (5);
             \draw[thick] (5) edge  (6);
               \draw[thick] (6) edge  (7);
          

               \draw[thick, dashed] (1) edge[bend left=45] (3);
            \draw[thick, dashed] (3) edge[bend right=45] (1);

               \draw[thick, dashed] (3) edge[bend left=45] (5);
           \draw[thick, dashed] (5) edge[bend right=45] (3);          
            
            \draw[thick, dashed] (5) edge[bend left=45] (7);
            \draw[thick, dashed] (7) edge[bend right=45] (5);

        \end{tikzpicture}

%% file: tikzfigures/chain_bn.tikz
\begin{tikzpicture}[->,>=stealth',shorten >=1pt,auto,node distance=1.65cm,
  main node/.style={circle,draw,minimum size=1cm,font=\sffamily\small}]
        
            \node[main node, color=blue] (1) { $V_0$};
            \node[main node] (2)[right of =1] { $V_1$};
             \node[main node ](8) [above of =2] {$U_0$};

            \node[main node] (3)[right of =2] { $V_2$};
            \node[main node] (4)[right of=3]{$V_3$};
            \node[main node ](9) [above of =4] {$U_1$};

            \node[main node] (5)[right of=4]{$V_4$};         
            \node[main node] (6)[right of=5]{$V_5$};
            \node[main node ](10) [above of =6] {$U_2$};

            \node[main node, color=red] (7)[right of=6]{$V_6$};
           
                \draw[thick] (1) edge  (2);
               \draw[thick] (2) edge  (3);
                \draw[thick] (3) edge  (4);
               \draw[thick] (4) edge  (5);
             \draw[thick] (5) edge  (6);
               \draw[thick] (6) edge  (7);
                \draw[thick] (8) edge[bend right=25]  (1);
                \draw[thick] (8) edge[bend left=25]   (3);
                \draw[thick] (9) edge[bend right=25]   (3);
                \draw[thick] (9) edge[bend left=25]   (5);
                \draw[thick] (10) edge[bend right=25]   (5);
                \draw[thick] (10) edge[bend left=25]   (7);

        \end{tikzpicture}

%% file: tikzfigures/chain_trunc.tikz
\begin{tikzpicture}[->,>=stealth',shorten >=1pt,auto,node distance=1.65cm,
  main node/.style={circle,draw,minimum size=1cm,font=\sffamily\small}]
        
            \node[main node, color =blue] (1) { $V_0$};
            \node[main node] (2)[right of =1] { $V_1$};
             \node[main node ](8) [above of =2] {$U_0$};

            \node[main node] (3)[right of =2] { $V_2$};
            \node[main node] (4)[right of=3]{$V_3$};
            \node[main node ](9) [above of =4] {$U_1$};

            \node[main node] (5)[right of=4]{$V_4$};         
            \node[main node] (6)[right of=5]{$V_5$};
            \node[main node ](10) [above of =6] {$U_2$};

            \node[main node, color=red] (7)[right of=6]{$V_6$};
           
                \draw[thick] (1) edge  (2);
               \draw[thick] (2) edge  (3);
                \draw[thick] (3) edge  (4);
               \draw[thick] (4) edge  (5);
             \draw[thick] (5) edge  (6);
               \draw[thick] (6) edge  (7);
                \draw[thick] (8) edge[bend left=25]   (3);
                \draw[thick] (9) edge[bend right=25]   (3);
                \draw[thick] (9) edge[bend left=25]   (5);
                \draw[thick] (10) edge[bend right=25]   (5);
                \draw[thick] (10) edge[bend left=25]   (7);

        \end{tikzpicture}

%% file: tikzfigures/model1.tikz
\begin{tikzpicture}[->,>=stealth',shorten >=1pt,auto,node distance=1.5cm,
  main node/.style={circle,draw,font=\sffamily\small}]

  \node[main node] (1) {W};
  \node[main node] (2) [below left of=1] {R};
  \node[main node, color=blue] (3) [below left of =2] {X};
  \node[main node, color=red] (4) [right of=3] {Y};
  
   \draw[thick] (1) edge  (2);
  \draw[thick, dashed] (1) edge[bend right=25] (3);
    \draw[thick, dashed] (3) edge[bend left=25] (1);

    \draw[thick] (2) edge  (3);
    \draw[thick] (3) edge  (4);

  \draw[thick, dashed] (1) edge[bend left=25] (4);
    \draw[thick, dashed] (4) edge[bend right=25] (1);

\end{tikzpicture}

%% file: tikzfigures/model3.tikz
 \begin{tikzpicture}[->,>=stealth',shorten >=1pt,auto,node distance=1.5cm,
  main node/.style={circle,draw,font=\sffamily\small}]

  \node[main node] (1) {W};
  \node[main node] (4) [below of=1] {Z};
  \node[main node, color=blue] (2)[ left of=4] {X};
  \node[main node, color = red] (3) [ right of=4] {Y};

   \draw[thick] (1) edge  (4);
   \draw[thick] (1) edge  (2);
    \draw[thick] (1) edge  (3);
   \draw[thick] (3) edge  (4);
   \draw[thick] (2) edge  (4);

  \draw[thick, dashed] (1) edge[bend right=25] (2);
    \draw[thick, dashed] (2) edge[bend left=25] (1);
      \draw[thick, dashed] (1) edge[bend left=25] (3);
    \draw[thick, dashed] (3) edge[bend right=25] (1);
      \draw[thick, dashed] (2) edge[bend right=45] (3);
    \draw[thick, dashed] (3) edge[bend left=45] (2);
   
   \end{tikzpicture}

%% file: tikzfigures/model8.tikz
 \begin{tikzpicture}[->,>=stealth',shorten >=1pt,auto,node distance=1.5cm,
  main node/.style={circle,draw,font=\sffamily\small}]

  \node[main node] (1) {W};
 \node[main node, color=blue] (4) [below right of=1] {R};
  \node[main node, color = red] (5) [ below right of=4] {Y};
    \node[main node] (2)[  left of=5] {Z};
 \node[main node,  color=blue] (3) [left of=2] {X};

   \draw[thick] (1) edge  (4);
   \draw[thick] (1) edge  (2);
    \draw[thick] (1) edge  (3);
   \draw[thick] (3) edge  (2);
   \draw[thick] (2) edge  (5);
   \draw[thick] (4) edge  (5);

  \draw[thick, dashed] (1) edge[bend left=25] (5);
    \draw[thick, dashed] (5) edge[bend right=25] (1);
    
      \draw[thick, dashed] (1) edge[bend right=25] (3);
    \draw[thick, dashed] (3) edge[bend left=25] (1);
    
      \draw[thick, dashed] (5) edge[bend left=45] (3);
    \draw[thick, dashed] (3) edge[bend right=45] (5);
   
   \end{tikzpicture}

%% file: tikzfigures/diamond.tikz
\begin{tikzpicture}[->,>=stealth',shorten >=1pt,auto,node distance=1.25cm,
  main node/.style={ellipse,minimum size=.5pt ,draw,font=\sffamily\small}]
            \node[main node, color =blue] (0) { $V_0$};
            \node[main node] (1)[above of =0] { $V_1$};
            \node[main node] (2)[above of =1] { $V_2$};
            \node[main node] (3) [above of =2] {$V_3$};
            \node[main node] (4) [above of =3] {$V_4$};
            \node[main node] (5)[right of =0] { $V_5$};
            \node[main node] (6)[right of =5] { $V_6$};
            \node[main node] (7) [right of =6] {$V_7$};
            \node[main node] (8) [right of =7] {$V_8$};
            \node[main node] (9)[left of =0] { $V_9$};
            \node[main node] (10)[left of =9] { $V_{10}$};
            \node[main node] (11) [left of =10] {$V_{11}$};
            \node[main node] (12) [left of =11] {$V_{12}$};
           \node[main node] (13)[below of=0]{$V_{13}$};
            \node[main node] (14)[below of =13] { $V_{14}$};
            \node[main node] (15) [below of =14] {$V_{15}$};
           \node[main node, color=red] (16) [below of =15] {$V_{16}$};

                \draw[thick] (0) edge  (1);
                \draw[thick] (0) edge  (5);
                \draw[thick] (0) edge  (9);
                \draw[thick] (0) edge  (13);

                \draw[thick] (1) edge  (2);
               \draw[thick] (2) edge  (3);
                \draw[thick] (3) edge  (4);
               \draw[thick] (5) edge  (6);
               \draw[thick] (6) edge  (7);
               \draw[thick] (7) edge  (8);

               \draw[thick] (9) edge  (10);
               \draw[thick] (10) edge  (11);
               \draw[thick] (11) edge  (12);
               
               \draw[thick] (13) edge  (14);
               \draw[thick] (14) edge  (15);
               \draw[thick] (15) edge  (16);

                \draw[thick] (8) edge  (16);
               \draw[thick] (12) edge  (16);
               \draw[thick] (4) edge [bend left=40]  (16);

               \draw[thick, dashed] (0) edge[bend left=45] (2);
            \draw[thick, dashed] (2) edge[bend right=45] (0);
            
            \draw[thick, dashed] (2) edge[bend left=45] (4);
            \draw[thick, dashed] (4) edge[bend right=45] (2);

            \draw[thick, dashed] (0) edge[bend left=45] (6);
            \draw[thick, dashed] (6)edge[bend right=45] (0);  
            
            \draw[thick, dashed] (6) edge[bend left=45] (8);
            \draw[thick, dashed] (8)edge[bend right=45] (6);        
            
            \draw[thick, dashed] (0) edge[bend left=45] (10);
            \draw[thick, dashed] (10) edge[bend right=45] (0);  
            
            \draw[thick, dashed] (10) edge[bend left=45] (12);
            \draw[thick, dashed] (12) edge[bend right=45] (10); 
            
            \draw[thick, dashed] (0) edge[bend right=45] (14);
            \draw[thick, dashed] (14) edge[bend left=45] (0);
            \draw[thick, dashed] (14) edge[bend right=45] (16);
            \draw[thick, dashed] (16) edge[bend left=45] (14);

        \end{tikzpicture}

%% file: tikzfigures/cone_cloud.tikz
\begin{tikzpicture}[->,>=stealth',shorten >=1pt,auto,node distance=1.25cm,
  main node/.style={ellipse,minimum size=.5pt ,draw,font=\sffamily\small}]
            \node[main node, color =red] (0) { $V_0$};
            \node[main node] (1)[above left of =0] { $V_1$};
            \node[main node] (2)[above right of =0] { $V_2$};
            \node[main node] (3) [above left of =1] {$V_3$};
            \node[main node,  color =blue] (4) [above right of =1] {$V_4$};
            \node[main node] (5)[above right of =2] { $V_5$};
            \node[main node] (6)[above left of =3] { $V_6$};
            \node[main node] (7) [above right of =3] {$V_7$};
            \node[main node] (8) [above right of =4] {$V_8$};
            \node[main node] (9)[above right of =5] { $V_9$};
            \node[main node,  color =blue] (10)[above left of =6] { $V_{10}$};
            \node[main node] (11) [above right of =6] {$V_{11}$};
            \node[main node] (12) [above right of =7] {$V_{12}$};
           \node[main node] (13)[above right  of=8]{$V_{13}$};
            \node[main node,  color =blue] (14)[above right of =9] { $V_{14}$};

                \draw[thick] (1) edge  (0);
                \draw[thick] (2) edge  (0);
                \draw[thick] (3) edge  (1);
                \draw[thick] (4) edge  (1);
                \draw[thick] (4) edge  (2);
               \draw[thick] (5) edge  (2);
                \draw[thick] (6) edge  (3);
               \draw[thick] (7) edge  (3);
               \draw[thick] (7) edge  (4);
               \draw[thick] (8) edge  (4);

               \draw[thick] (8) edge  (5);
               \draw[thick] (9) edge  (5);
               \draw[thick] (10) edge  (6);
               
               \draw[thick] (11) edge  (6);
               \draw[thick] (11) edge  (7);
               \draw[thick] (12) edge  (7);

                \draw[thick] (12) edge  (8);
               \draw[thick] (13) edge  (8);
                \draw[thick] (13) edge  (9);
               \draw[thick] (14) edge  (9);

               \draw[thick, dashed] (0) edge[bend right=45] (5);
            \draw[thick, dashed] (5) edge[bend left=45] (0);
            
            \draw[thick, dashed] (5) edge[bend right=45] (14);
            \draw[thick, dashed] (14) edge[bend left=45] (5);

            \draw[thick, dashed] (0) edge[bend left=45] (3);
            \draw[thick, dashed] (3)edge[bend right=45] (0);  
            
            \draw[thick, dashed] (3) edge[bend left=45] (10);
            \draw[thick, dashed] (10)edge[bend right=45] (3);        
            
            \draw[thick, dashed] (11) edge[bend right=45] (10);
            \draw[thick, dashed] (10) edge[bend left=45] (11);  
            
            \draw[thick, dashed] (11) edge[bend left=45] (12);
            \draw[thick, dashed] (12) edge[bend right=45] (11); 
            
            \draw[thick, dashed] (12) edge[bend left=45] (13);
            \draw[thick, dashed] (13) edge[bend right=45] (12);
            
            \draw[thick, dashed] (14) edge[bend right=45] (13);
            \draw[thick, dashed] (13) edge[bend left=45] (14);

        \end{tikzpicture}

%% file: tikzfigures/chain_9.tikz
\begin{tikzpicture}[->,>=stealth',shorten >=1pt,auto,node distance=1.65cm,
  main node/.style={circle,draw,minimum size=1cm,font=\sffamily\small}]
        
            \node[main node, color =blue] (1) { $V_0$};
            \node[main node] (2)[right of =1] { $V_1$};
            \node[main node] (3)[right of =2] { $V_2$};
            \node[main node] (4)[right of=3]{$V_3$};
            \node[main node] (5)[right of=4]{$V_4$};         
            \node[main node] (6)[right of=5]{$V_5$};
            \node[main node] (7)[right of=6]{$V_6$};
            \node[main node] (8)[right of=7]{$V_7$};
            \node[main node, color=red] (9)[right of=8]{$V_8$};

                \draw[thick] (1) edge  (2);
               \draw[thick] (2) edge  (3);
                \draw[thick] (3) edge  (4);
               \draw[thick] (4) edge  (5);
             \draw[thick] (5) edge  (6);
               \draw[thick] (6) edge  (7);
             \draw[thick] (7) edge  (8);
             \draw[thick] (8) edge  (9);

               \draw[thick, dashed] (1) edge[bend left=45] (3);
            \draw[thick, dashed] (3) edge[bend right=45] (1);

               \draw[thick, dashed] (3) edge[bend left=45] (5);
            \draw[thick, dashed] (5) edge[bend right=45] (3);          
            
            \draw[thick, dashed] (5) edge[bend left=45] (7);
            \draw[thick, dashed] (7) edge[bend right=45] (5);    
        
             \draw[thick, dashed] (7) edge[bend left=45] (9);
            \draw[thick, dashed] (9) edge[bend right=45] (7);      

        \end{tikzpicture}

%% file: tikzfigures/3graphs.tikz
\begin{tikzpicture}

\node[anchor=south] at (0,.8) {\rotatebox{90}{\tiny Error (mad)}};
\node[anchor=south west] at (0,0) {\includegraphics[width=4cm]{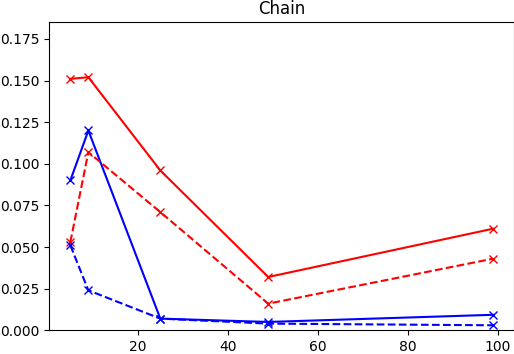}};
\node[anchor=south] at (6,-.25) {{\tiny Number of Variables }};

\node[anchor=south west] at (4.2,0) {\includegraphics[width=3.65cm]{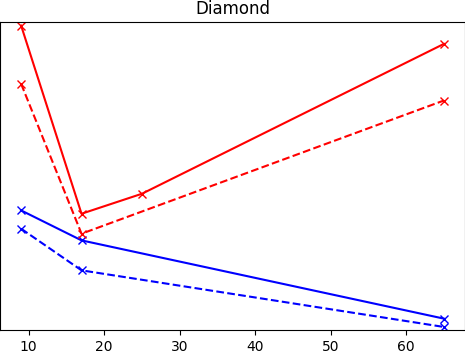}};

\node[anchor=south west] at (8.2,0) {\includegraphics[width=3.65cm]{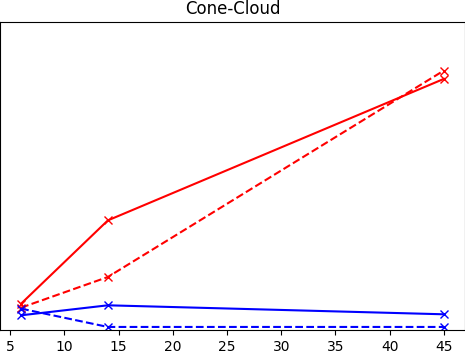}};
\node[anchor=north west] at (12,3) {\includegraphics[width=2cm]{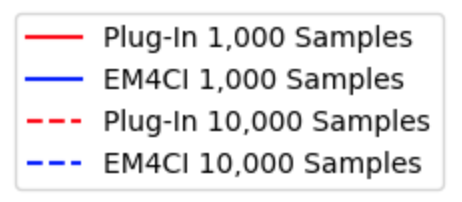}};

\end{tikzpicture}

%% file: tikzfigures/model2.tikz
  \begin{tikzpicture}[->,>=stealth',shorten >=1pt,auto,node distance=1.5cm,
  main node/.style={circle,draw,font=\sffamily\small}]

  \node[main node, color=blue] (1) {$X_1$};
  \node[main node, color=blue] (2) [below of=1] {$X_2$};
  \node[main node] (3) [below left of=1] {R};
  \node[main node, color=red] (4) [below left of=3] {Y};
  \node[main node] (5) [right of=2] {Z};
  
   \draw[thick] (1) edge  (2);
   \draw[thick] (1) edge  (3);
   \draw[thick] (3) edge  (4);
  \draw[thick, dashed] (1) edge[bend right=25] (4);
    \draw[thick, dashed] (4) edge[bend left=25] (1);

    \draw[thick] (2) edge  (5);
    \draw[thick] (1) edge  (5);
    \draw[thick] (2) edge  (4);
    
  \draw[thick, dashed] (1) edge[bend left=25] (5);
    \draw[thick, dashed] (5) edge[bend right=25] (1);
  \draw[thick, dashed] (4) edge[bend right=25] (5);
    \draw[thick, dashed] (5) edge[bend left=25] (4);
   \end{tikzpicture}

%% file: tikzfigures/model4.tikz
 \begin{tikzpicture}[->,>=stealth',shorten >=1pt,auto,node distance=1.5cm,
  main node/.style={circle,draw,font=\sffamily\small}]

  \node[main node, color=blue] (1) {$X_1$};
  \node[main node] (5) [ below left of=1] {R}; 
  \node[main node, color=red] (6) [ below left of=5] {Y};
  \node[main node] (7) [ below right of=6] {S}; 
    \node[main node, color=blue] (2) [below right of=5] {$X_2$};
  \node[main node] (4) [ right of=2] {Z};
    \node[main node, color=blue] (3)[ below right of=7] {$X_3$};

   \draw[thick] (1) edge  (5);
   \draw[thick] (1) edge  (2);
   \draw[thick] (2) edge  (3);
   \draw[thick] (5) edge  (6);
    \draw[thick] (1) edge  (4);
   \draw[thick] (4) edge  (3);
   \draw[thick] (4) edge  (2);
   \draw[thick] (2) edge  (6);
   \draw[thick] (7) edge  (6);
   \draw[thick] (3) edge  (7);
   
  \draw[thick, dashed] (1) edge[bend right=25] (6);
    \draw[thick, dashed] (6) edge[bend left=25] (1);
    
      \draw[thick, dashed] (1) edge[bend left=25] (4);
    \draw[thick, dashed] (4) edge[bend right=25] (1);
    
      \draw[thick, dashed] (3) edge[bend right=25] (4);
    \draw[thick, dashed] (4) edge[bend left=25] (3);

      \draw[thick, dashed] (3) edge[bend left=25] (6);
    \draw[thick, dashed] (6) edge[bend right=25] (3);
   \end{tikzpicture}

%% file: tikzfigures/model5.tikz
\begin{tikzpicture}[->,>=stealth',shorten >=1pt,auto,node distance=1.5cm,
  main node/.style={circle,draw,font=\sffamily\small}]

  \node[main node, color=blue] at (0,0) (1) {$X$};
  \node[main node] (2) [ below right of=1] {$Z_1$}; 
  
  \node[main node]  at (4,0) (3){$Z_2$}; 
  \node[main node] (4) [ below left of=3] {$Z_3$}; 
  \node[main node, color=red] (5) [ below left of=4] {Y};

   \draw[thick] (1) edge  (2);
   \draw[thick] (3) edge  (1);
   \draw[thick] (3) edge  (4);
   \draw[thick] (4) edge  (5);
    \draw[thick] (2) edge  (5);
    \draw[thick] (3) edge  (2);  
   
  \draw[thick, dashed] (1) edge[bend left=25] (3);
    \draw[thick, dashed] (3) edge[bend right=25] (1);
    
      \draw[thick, dashed] (1) edge[bend right=45] (5);
    \draw[thick, dashed] (5) edge[bend left=45] (1);
    
      \draw[thick, dashed] (3) edge[bend left=40] (5);
    \draw[thick, dashed] (5) edge[bend right=40] (3);

      \draw[thick, dashed] (1) edge[bend left=5] (4);
    \draw[thick, dashed] (4) edge[bend right=5] (1);
   \end{tikzpicture}

%% file: tikzfigures/model6.tikz
  \begin{tikzpicture}[->,>=stealth',shorten >=1pt,auto,node distance=1.5cm,
  main node/.style={circle,draw,font=\sffamily\small}]

  \node[main node] (1) {$Z_3$};
  \node[main node] (2) [ below  of=1] {$Z_2$}; 
  \node[main node, color=blue] (3) [  left of=2] {$X_2$};
    \node[main node, color=red] (5) [below of=2] {$Y$};
  \node[main node] (6) [ left of=5] {$Z_1$}; 
  \node[main node, color=blue] (4) [  left of=6] {$X_1$};

   \draw[thick] (1) edge  (2);
   \draw[thick] (1) edge  (3);
   \draw[thick] (3) edge  (5);
    \draw[thick] (3) edge  (4);
   \draw[thick] (4) edge  (6);
   \draw[thick] (6) edge  (5);
   \draw[thick] (2) edge  (5);

  \draw[thick, dashed] (2) edge[bend right=25] (3);
    \draw[thick, dashed] (3) edge[bend left=25] (2);

      \draw[thick, dashed] (3) edge[bend right=25] (4);
    \draw[thick, dashed] (4) edge[bend left=25] (3);

   \end{tikzpicture}

%% file: tikzfigures/model7.tikz
  \begin{tikzpicture}[->,>=stealth',shorten >=1pt,auto,node distance=1.5cm,
  main node/.style={circle,draw,font=\sffamily\small}]

  \node[main node] (1) {$Z_3$};
  \node[main node] (2) [ below  of=1] {$Z_2$}; 
  \node[main node, color=blue] (3) [  left of=2] {$X_2$};
    \node[main node, color=red] (5) [below of=2] {$Y$};
  \node[main node] (6) [ left of=5] {$Z_1$}; 
  \node[main node, color=blue] (4) [  left of=6] {$X_1$};

   \draw[thick] (1) edge  (2);
   \draw[thick] (1) edge  (3);
   \draw[thick] (3) edge  (5);
    \draw[thick] (3) edge  (4);
   \draw[thick] (4) edge  (6);
   \draw[thick] (6) edge  (5);
   \draw[thick] (2) edge  (5);

  \draw[thick, dashed] (2) edge[bend right=25] (3);
    \draw[thick, dashed] (3) edge[bend left=25] (2);

      \draw[thick, dashed] (3) edge[bend right=25] (4);
    \draw[thick, dashed] (4) edge[bend left=25] (3);

      \draw[thick, dashed] (6) edge[bend right=60] (1);
    \draw[thick, dashed] (1) edge[bend left=60] (6);

   \end{tikzpicture}

%% file: tikzfigures/model9.tikz
 \begin{tikzpicture}[->,>=stealth',shorten >=1pt,auto,node distance=1.5cm,
  main node/.style={circle,draw,font=\sffamily\small}]

  \node[main node] (1) {W};
  \node[main node] (4) [below of=1] {Z};
  \node[main node, color=blue] (2)[ left of=4] {X};
  \node[main node, color = red] (3) [ right of=4] {Y};

   \draw[thick] (1) edge  (4);
   \draw[thick] (1) edge  (2);
    \draw[thick] (1) edge  (3);
   \draw[thick] (4) edge  (3);
   \draw[thick] (2) edge  (4);

  \draw[thick, dashed] (1) edge[bend right=25] (2);
    \draw[thick, dashed] (2) edge[bend left=25] (1);
      \draw[thick, dashed] (1) edge[bend left=25] (3);
    \draw[thick, dashed] (3) edge[bend right=25] (1);
      \draw[thick, dashed] (2) edge[bend right=45] (3);
    \draw[thick, dashed] (3) edge[bend left=45] (2);
   
   \end{tikzpicture}

%% file: tikzfigures/chain.tikz
\begin{tikzpicture}[->,>=stealth',shorten >=1pt,auto,node distance=1.65cm,
  main node/.style={circle,draw,minimum size=1cm,font=\sffamily\small}]
        
            \node[main node, color =blue] (1) { $Z_1$};
            \node[main node] (2)[right of =1] { $Z_2$};
            \node[main node] (3)[right of =2] { $Z_3$};
            \node[main node] (4)[right of=3]{$Z_4$};
               \node[main node] (5)[right of=4]{$Z_5$};         
            \node[] (6)[right of=5]{................};

             \node[main node] (9)[right of=6]{$Z_{n-2}$};
           \node[main node] (10)[right of=9]{$Z_{n-1}$};
            \node[main node, color=red] (11)[right of=10]{$Z_n$};
            \node[] (7)[above right of=5]{.};
            \node[] (8)[above left of=9]{.};

                \draw[thick] (1) edge  (2);
               \draw[thick] (2) edge  (3);
                \draw[thick] (3) edge  (4);
               \draw[thick] (4) edge  (5);
             \draw[thick] (5) edge  (6);
               \draw[thick] (6) edge  (9);
               \draw[thick] (9) edge  (10);
               \draw[thick] (10) edge  (11);
               
               \draw[thick, dashed] (1) edge[bend left=45] (3);
            \draw[thick, dashed] (3) edge[bend right=45] (1);

               \draw[thick, dashed] (3) edge[bend left=45] (5);
            \draw[thick, dashed] (5) edge[bend right=45] (3);          
            
            \draw[thick, dashed] (9) edge[bend left=45] (11);
            \draw[thick, dashed] (11) edge[bend right=45] (9);    

            \draw[thick, dashed] (7) edge[bend right=25] (5);
            \draw[thick, dashed] (8) edge[bend left=25] (9);


        \end{tikzpicture}